\documentclass[10pt,twocolumn,letterpaper]{article}

\usepackage[
  letterpaper,
  margin=1in,
  columnsep=0.25in
]{geometry}

\usepackage[T1]{fontenc}
\usepackage{newtxtext}
\usepackage{helvet}
\usepackage{courier}
\usepackage{microtype}
\usepackage[hyphens]{url}
\usepackage{graphicx}
\graphicspath{{Paper_figures/}{supplementary_figures/}}
\usepackage{natbib}
\setcitestyle{authoryear,round,aysep={}}

\usepackage{amsmath,amsfonts}
\usepackage{booktabs}
\usepackage{tabularx}
\usepackage{array}
\usepackage{caption}
\usepackage{placeins}

\usepackage[hidelinks]{hyperref}
\hypersetup{
  pdftitle={Verifier-Guided Model Discovery for Physical Dynamical Systems with Pretrained Symbolic Transformers},
  pdfauthor={Farbod Faraji and Francesco Belardinelli}
}

\usepackage{titling}

\makeatletter
\renewcommand\section{\@startsection{section}{1}{\z@}
  {-2.0ex plus -0.5ex minus -0.2ex}
  {0.8ex plus 0.2ex}
  {\large\bfseries}}

\renewcommand\subsection{\@startsection{subsection}{2}{\z@}
  {-2.0ex plus -0.5ex minus -0.2ex}
  {0.8ex plus 0.2ex}
  {\normalsize\bfseries}}

\renewcommand\paragraph{\@startsection{paragraph}{4}{\z@}
  {-6pt plus -2pt minus -1pt}
  {-1em}
  {\normalsize\bfseries}}
\makeatother

\pretitle{\centering\LARGE\bfseries}
\posttitle{\par\vspace{0.6em}}

\preauthor{\centering\normalsize}
\postauthor{\par\vspace{-1.5em}}

\newcommand{\Rey}{\mathrm{Re}}

\title{Verifier-Guided Model Discovery for Physical Dynamical Systems\\
with Pretrained Symbolic Transformers}

\author{%
Farbod Faraji \quad Francesco Belardinelli\\[0.45em]
\small Department of Computing, Huxley Building, Imperial College London\\
\small 180 Queen's Gate, South Kensington, London SW7 2RH, United Kingdom
}

\date{}

\begin{document}

\maketitle

\begin{abstract}
Reliable forecasting of nonlinear physical systems underpins scientific discovery and engineering decision-making. Yet high-fidelity simulations are prohibitively costly, and machine-learning surrogates can be opaque and encode assumptions about system dynamics, limiting generalizability. Pretrained transformers mapping synthetic ODE trajectories to equations offer interpretable alternatives, promising transfer without system-specific equation knowledge. Transferring them reliably to high-dimensional physical data, however, remains an open challenge. We develop a verifier-guided (VG) workflow around ODEFormer as a symbolic backbone, using dynamical and physical-admissibility criteria to select from a multi-trajectory candidate equation pool, enabling transfer. On canonical Van der Pol oscillators, VG outperforms the original ODEFormer workflow across held-out initial conditions. We then address vortex shedding---a phenomenon occurring in atmospheric and plasma systems of societal relevance---through coordinate reduction and symbolic discovery at fixed and varying Reynolds numbers. VG discovers fixed-parameter reduced-order equations that recover the fundamental shedding oscillator and higher harmonics without a wake-specific candidate library or prescribed Navier--Stokes structure, while the cross-parameter model generalizes to withheld regimes. Reconstruction fidelity alone did not determine symbolic discoverability, highlighting the importance of compatibility between latent dynamics and the backbone's pretraining distribution. This work establishes a verifier-guided neural-to-symbolic methodology for interpretable and physically auditable forecasting in the natural sciences.
\end{abstract}

\section{Introduction}

In order to predict the evolution of nonlinear physical systems, computational efficiency alone is insufficient when the predictions of a model cannot be interpreted or audited. For AI in the natural sciences, forecasts should be accompanied by evidence that the inferred dynamics are physically credible and by clearly identified limits. Symbolic reduced-order models support such scrutiny: their explicit equations can be examined against dynamical and physical requirements \citep{faraji2025perspective}, while remaining efficient enough for parameter exploration, design optimization, and control.

Pretrained symbolic transformers map trajectories to such explicit models without case-specific candidate libraries. Their reliable transfer from synthetic pretraining to physical systems, however, requires more than generating equations reproducing individual trajectories. Pretraining uses synthetic equation--trajectory pairs drawn from a prescribed symbolic vocabulary, while reduced physical coordinates may omit unresolved dynamics needed for a closed autonomous description. Generated equations may still be unstable, fail across initial conditions or parameters, or violate physical requirements. We address this challenge with a verifier-guided (VG) workflow that uses executable tests of dynamical behavior and physical admissibility to select equations across multiple trajectories.

Testing this approach beyond directly observed ODEs requires a high-dimensional physical problem. Vortex shedding provides such a setting with broad scientific and societal relevance: atmospheric von K\'arm\'an vortex streets occur in island wakes \citep{jma2019,etling1990}; related instabilities drive vortex-induced loading \citep{williamson2004} and arise in plasma flows past obstacles \citep{gruszecki2010,bailung2020}. The controlled cylinder-flow case examined in this work thus connects auditable reduced-order discovery to atmospheric observation, aerospace and transportation, infrastructure safety, and space-plasma environments.

The principal contributions of this work are:

\begin{itemize}
\item \textbf{Verifier-guided symbolic discovery.} We place executable tests of dynamical behavior and physical admissibility at the center of multi-trajectory model selection using a pretrained symbolic transformer.

\item \textbf{Generalization across initial conditions.} On fixed Van der Pol oscillators, VG outperforms the baseline single-trajectory transformer pipeline for every held-out initial condition while closely recovering the governing structure.

\item \textbf{Transfer to high-dimensional physical data.} At fixed and varying Reynolds numbers, VG recovers shedding dynamics and yields stable predictions at withheld interpolation and extrapolation values; representation fidelity alone does not determine symbolic discoverability.
\end{itemize}

\section{Related Work}

\paragraph{Predictive and physics-informed dynamics learning.}
Data-driven modeling of dynamical systems offers several routes to efficient forecasting. Optimized dynamic mode decomposition (OPT-DMD) identifies coherent spatiotemporal modes and their best-fit linear evolution, reducing noise-induced bias and, when constrained, enabling stable forecasting \citep{askham2018,faraji2024}. Neural ODEs parameterize continuous-time vector fields, while neural operators, including the Fourier Neural Operator, learn mappings between function spaces for families of PDE solutions \citep{chen2018,li2021}. These approaches do not generally recover explicit nonlinear governing laws: OPT-DMD uses linear evolution in its coordinates, while neural ODEs and neural operators retain opaque parameterizations. Physics-informed neural networks incorporate known governing-equation residuals and boundary or initial conditions, whereas structure-preserving architectures encode properties such as Hamiltonian conservation \citep{raissi2019,greydanus2019}. Such methods require the relevant physical structure to be known and supplied in advance. Residual penalties promote but do not guarantee constraint satisfaction; hard constraints can enforce known, expressible requirements \citep{lu2021}.

\paragraph{Sparse-regression approaches to symbolic discovery.}
Symbolic dynamics discovery instead seeks explicit governing equations. Sparse Identification of Nonlinear Dynamics (SINDy) identifies parsimonious systems through sparse regression over a prescribed candidate library \citep{brunton2016}. Weak-form, ensemble, and Bayesian variants reduce sensitivity to numerical differentiation, improve robustness to limited or noisy data, and quantify model uncertainty \citep{messenger2021,fasel2022,fung2025}. Physical structure can also enter regression: constrained sparse Galerkin regression combines POD reduction with energy-preserving constraints in fluid reduced-order models \citep{loiseau2018}. Phi Method identifies discretized evolution operators from a candidate library, avoiding explicit derivatives and continuous-time integration \citep{faraji2025}. AutoSINDy \citep{basiri2026} combines PySR candidate generation \citep{cranmer2023}, library curation, and sparse identification to mitigate the difficulty of prescribing an expressive library.

\paragraph{Search-based and generative symbolic dynamics discovery.}
Many general-purpose search-based and neural generative symbolic regression methods were initially developed for static relations, $y=f(\mathbf{x})$, with later extensions to dynamics. DynAIFeynman applies AI Feynman's separability, symmetry, and dimensional-consistency tests to states paired with finite-difference derivatives, while ProGED samples grammar-defined structures and fits their parameters against estimated derivatives or by simulating candidate ODEs \citep{udrescu2020,weilbach2021,brence2021,omejc2024}. Neural generators learn distributions over expressions: Deep Symbolic Regression uses a recurrent generator with risk-seeking policy gradients, while pretrained transformer approaches include NeSymReS, SymbolicGPT, and end-to-end symbolic regression; TPSR further adds Monte Carlo tree search with accuracy and complexity feedback \citep{petersen2021,biggio2021,valipour2021,kamienny2022,shojaee2023}. ODEFormer specializes this formulation to dynamics discovery by mapping multivariate trajectories directly to coupled ODE systems \citep{dascoli2024}. MIO trains a trajectory-to-equation transformer to infer shared dynamics from multiple trajectories of the same system \citep{sahin2025}. Diffusion-based symbolic regression provides a non-autoregressive alternative through masked iterative denoising and dataset-specific reinforcement learning \citep{bastiani2025}. These approaches expand or accelerate symbolic search but retain representational choices through observed variables, primitives, grammar, objectives, or pretraining.

\paragraph{Toward knowledge-guided and verified discovery.}
Recent work increasingly makes scientific knowledge operational during equation discovery. LLM-assisted physics-informed symbolic regression adds language-model assessments to the search objective; prior-guided methods use executable constraint programs to steer evolutionary search; and Latent Grammar Flow embeds stability constraints in grammar rules or conditions generation on them \citep{taskin2026,xiao2026,yu2026}. These approaches place scientific consistency within search or generation. Our complementary objective is to transfer an existing pretrained symbolic transformer without modifying or retraining its generator. We leave our chosen pretrained backbone, ODEFormer \citep{dascoli2024}, unchanged and use executable dynamical and physical-admissibility tests on pooled candidates generated independently across trajectories to select a common equation system. This enables transfer from synthetic pretraining to reduced physical coordinates while retaining symbolic transparency and avoiding a case-specific candidate library or prescribed governing structure.

\section{Background and Problem Formulation}

Consider $M$ observed trajectories $\mathbf{x}^{(m)}(t)$, where $m$ indexes different initial conditions, time windows, or parameter values. Symbolic dynamics discovery seeks an explicit model
\begin{equation}
\label{eq:general-system}
\dot{\mathbf{x}}=\mathbf{f}(\mathbf{x}),
\qquad
\mathbf{f}:\mathcal{X}\subseteq\mathbb{R}^{d}\rightarrow\mathbb{R}^{d},
\end{equation}
whose initial-value problems yield finite trajectories over the state domain and time horizons of interest. Equation~\eqref{eq:general-system} treats the supplied coordinates as a sufficient state for a deterministic, autonomous, first-order description. We do not prescribe a case-specific parametric form or candidate library for $\mathbf{f}$ in this work; the representable candidates are nevertheless shaped by the learned distribution of the symbolic backbone.

We use ODEFormer as this backbone because it maps multivariate trajectories directly to coupled symbolic ODEs without numerical derivative targets, case-specific retraining, or a manually constructed function library \citep{dascoli2024}. Its encoder processes tokenized time--state observations, and its autoregressive decoder constructs the right-hand sides of the ODE system as symbolic sequences in prefix notation. ODEFormer thus induces an implicit structural prior through its synthetic pretraining distribution, including the symbolic vocabulary, expression-tree complexity, coefficient and trajectory distributions, and training-time filtering of unstable or uninformative trajectories. At inference, beam sampling produces multiple candidate equation systems \citep{dascoli2024}. The beam size controls how many candidate sequences are explored, while the temperature controls the concentration of the token distribution and thus candidate diversity. We use a beam size of $20$ and temperature of $0.1$; their sensitivity is examined on the canonical Van der Pol problem in Technical Appendix~\ref{sec:supp-beam-temperature}. Whereas the original ODEFormer workflow selects candidates by reconstruction of a single input trajectory, we reconsider how the generated candidates are evaluated and selected (Section~\ref{sec:method}).

For a spatially distributed physical state $\boldsymbol{\omega}(t)\in\mathbb{R}^{N}$, our objective is a symbolic ODE describing the temporal evolution of its dominant coordinates, rather than direct discovery of the underlying PDE. Let an encoder $\mathcal{E}$ and reconstruction map $\mathcal{R}$ define
\begin{equation}
\label{eq:reduced-system}
\mathbf{z}(t)=\mathcal{E}\!\left(\boldsymbol{\omega}(t)\right),
\qquad
\boldsymbol{\omega}(t)\approx\mathcal{R}\!\left(\mathbf{z}(t)\right),
\qquad
\dot{\mathbf{z}}=\mathbf{g}(\mathbf{z}),
\end{equation}
where $\mathbf{z}(t)\in\mathbb{R}^{d}$ and $d\ll N$. Eq.~\eqref{eq:reduced-system} provides a low-dimensional phase-space model whose equilibria, limit cycles, stability, frequencies, and modal couplings can be examined using established dynamical-systems tools. The primary role of coordinate reduction here is to define an interpretable state for the dominant temporal dynamics. The retained coordinates may omit higher-order dynamics important for closure; in that case, $\mathbf{g}$ represents an effective autonomous approximation within the chosen coordinates and the backbone's symbolic hypothesis class.

For trajectories observed at several values of a governing parameter $p$, we append the parameter to the state and constrain it to remain constant:
\begin{equation}
\label{eq:parametric-system}
\dot{\mathbf{z}}=\mathbf{g}(\mathbf{z},p),
\qquad
\dot{p}=0.
\end{equation}
The augmentation in Eq.~\eqref{eq:parametric-system} preserves the autonomous form expected by the backbone while allowing a single symbolic system to represent parameter-dependent dynamics. The resulting problem is to select an equation system that remains dynamically and physically admissible across multiple trajectories and at parameter values excluded from symbolic discovery.

\section{Methodology: Verifier-Guided Symbolic Model Discovery}
\label{sec:method}

\begin{figure}[t]
    \centering
    \includegraphics[width=0.8\columnwidth]{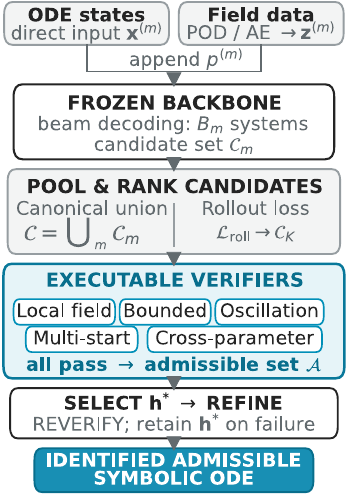}
    \caption{Verifier-guided symbolic model discovery. Independently decoded candidates are pooled, ranked, verified, and coefficient-refined before an admissible symbolic ODE is returned.}
    \label{fig:vg_workflow}
\end{figure}

Figure~\ref{fig:vg_workflow} illustrates the VG workflow. Directly observed ODE states enter unchanged, whereas high-dimensional fields are first reduced by POD or an autoencoder. The frozen backbone processes each trajectory or window separately; generated equations are then pooled, ranked, verified, and coefficient-refined.

\paragraph{Candidate generation, pooling, and ranking.}
Let $\boldsymbol{\xi}^{(m)}$ denote the trajectory or window supplied to the backbone: $\boldsymbol{\xi}^{(m)}=\mathbf{x}^{(m)}$ for observed ODE states, $\boldsymbol{\xi}^{(m)}=\mathbf{z}^{(m)}$ for reduced fields, and $\boldsymbol{\xi}^{(m)}=[\mathbf{z}^{(m)\top},p^{(m)}]^\top$ for cross-parameter discovery. For input $m$, the backbone retains $B_m$ decoded candidate systems indexed by $b$:
\begin{equation}
\label{eq:candidate_pool}
\mathcal{C}_m=\{\mathbf{h}_{m,b}\}_{b=1}^{B_m},
\qquad
\mathcal{C}=\bigcup_{m=1}^{M}\mathcal{C}_m.
\end{equation}
In Eq.~\eqref{eq:candidate_pool}, symbolic canonicalization removes duplicates before the union, which constitutes pooling.

Each $\mathbf{h}\in\mathcal{C}$ is ranked across discovery trajectories by Eq.~\eqref{eq:rollout-loss}:
\begin{equation}
\label{eq:rollout-loss}
\mathcal{L}_{\mathrm{roll}}(\mathbf{h})
=
\frac{1}{M}\sum_{m=1}^{M}
\frac{1}{d}\sum_{j=1}^{d}
\frac{
\operatorname{MSE}_{i}
\left(\widehat{\xi}_{ij}^{(m)},\xi_{ij}^{(m)}\right)
}{
\operatorname{Var}_{i}
\left(\xi_{ij}^{(m)}\right)+\epsilon
},
\end{equation}
where $d$ is the number of modeled dynamical coordinates, $i$ indexes time samples, and $\widehat{\boldsymbol{\xi}}^{(m)}$ is the candidate rollout initialized from the first observed state of trajectory $m$. The appended parameter is excluded from the $d$ coordinates in cross-parameter discovery. Nonfinite rollouts receive infinite loss, and the lowest-loss candidates form the shortlist $\mathcal{C}_K$; for cross-parameter discovery, the shortlist also retains the best candidates with explicit parameter dependence.

\paragraph{Executable verification and selection.}
The central component of VG is a suite of executable verifiers applied to $\mathcal{C}_K$. Each candidate defines the vector field $\dot{\boldsymbol{\xi}}=\mathbf{h}(\boldsymbol{\xi})$. Equation~\eqref{eq:local_verifier} measures its local agreement with observed dynamics:
\begin{equation}
V_{\mathrm{loc}}(\mathbf{h})
=
\frac{1}{d}\sum_{j=1}^{d}
\frac{
\operatorname{MSE}_{m,i}
\left(h_j(\boldsymbol{\xi}_{i}^{(m)}),
\dot{\xi}_{ij}^{(m)}\right)
}{
\operatorname{Var}_{m,i}
\left(\dot{\xi}_{ij}^{(m)}\right)+\epsilon
}
\label{eq:local_verifier}
\end{equation}
where $\dot{\xi}_{ij}^{(m)}$ is the numerically estimated rate of coordinate $j$ at observed sample $i$, and the error and variance are evaluated across discovery trajectories and times. A low score therefore indicates that the candidate vector field reproduces the locally observed directions and rates of state evolution.

Complementing this local test, rollout-based verifiers evaluate each candidate from prescribed starting states: observed states selected as independent initial conditions. They require finite and bounded evolution, agreement with observed oscillation amplitude and frequency or period, and consistency of these features across starting states. For cross-parameter systems, they additionally require a zero parameter equation, negligible integrated parameter drift, and explicit parameter dependence in at least one state equation.

Equation~\eqref{eq:admissible_selection} defines the admissible set and selects within it:
\begin{equation}
\begin{aligned}
\mathcal{A}
&=
\left\{
\mathbf{h}\in\mathcal{C}_K:
\mathsf{pass}_{\ell}(\mathbf{h})=1
\ \text{for every verifier }\ell
\right\},\\
\mathbf{h}^{\star}
&=
\underset{\mathbf{h}\in\mathcal{A}}{\arg\min}\,
\mathcal{L}_{\mathrm{roll}}(\mathbf{h}).
\end{aligned}
\label{eq:admissible_selection}
\end{equation}
Verification thus acts as an executable model-selection constraint rather than a post-hoc diagnostic. These tests establish empirical admissibility over prescribed domains and horizons. Technical Appendix~\ref{sec:supp-verifiers} reports their detailed definitions, thresholds, case-specific settings, and tolerance-sensitivity evaluation.

\paragraph{Identifiability and coefficient refinement.}
Viewed as an augmented observation map, the verifiers contract the set of trajectory-consistent candidates whenever they reject an otherwise indistinguishable model while retaining an admissible one. They can therefore improve identifiability within the generated candidate class; Technical Appendix~\ref{sec:supp-verifiers} formalizes this perspective and its local sensitivity-rank interpretation.

With the selected symbolic structure fixed, its decoded constants are jointly refined by Nelder--Mead minimization of $\mathcal{L}_{\mathrm{roll}}$ across the discovery trajectories. The refined equation is reverified; if it becomes inadmissible, VG returns the admissible unoptimized equation $\mathbf{h}^{\star}$.

\section{Empirical Evaluation}
\label{sec:results}

Evaluation considers the canonical Van der Pol (VdP) oscillator and high-dimensional flow past a stationary cylinder. VdP tests whether VG generalizes across unseen initial conditions, while the cylinder-flow problem tests transfer from synthetic ODE pretraining to reduced vortex-shedding dynamics at fixed and withheld values of Reynolds number.

\subsection{Canonical Dynamical System Case: Van der Pol Oscillator}
\label{sec:vdp-results}

The Van der Pol oscillator is a canonical self-excited system whose dynamics range from nearly harmonic motion to increasingly pronounced relaxation oscillations \citep{vanderpol1926}. Its first-order form is given in Eq.~\eqref{eq:vdp-system}:

\begin{equation}
\label{eq:vdp-system}
\begin{aligned}
\dot{x}_0 &= x_1,\\
\dot{x}_1 &= \mu(1-x_0^2)x_1-x_0\\
           &= \mu x_1-x_0-\mu x_0^2x_1.
\end{aligned}
\end{equation}

In Eq.~\eqref{eq:vdp-system}, $\dot{x}_0=x_1$ makes $x_1$ velocity-like; in $\dot{x}_1$, $-x_0$ restores while $\mu x_1$ injects energy near the origin and $-\mu x_0^2x_1$ dissipates it at larger amplitudes. Their balance produces a stable limit cycle. We consider \(\mu=0.5\) and \(1.5\) as separate fixed systems, representing weak and more pronounced nonlinear relaxation, respectively. This provides a controlled test of whether a discovered equation represents the surrounding vector field and transfers across initial conditions, rather than merely reproducing one observed orbit.

\paragraph{Experimental setup.}
At each value of \(\mu\), the original ODEFormer protocol \citep{dascoli2024} generated, ranked, and coefficient-optimized candidates from a single trajectory. We repeated this protocol for three preset initial conditions and report, for each test initial condition, the median across successful rollouts. VG instead decoded eight preset windows, ranked the pooled candidates over 24 windows from 12 preset initial conditions, verified the ten highest-ranked candidates, and applied the same coefficient optimization. Both protocols were evaluated on the same eight unseen initial conditions. Candidate-pool and test initial conditions were sampled independently from \([-1,1]^2\) using fixed seeds and were disjoint; Technical Appendix~\ref{sec:supp-vdp} reports the exact assignments.

\begin{figure}[t]
    \centering
    \includegraphics[width=0.95\columnwidth]{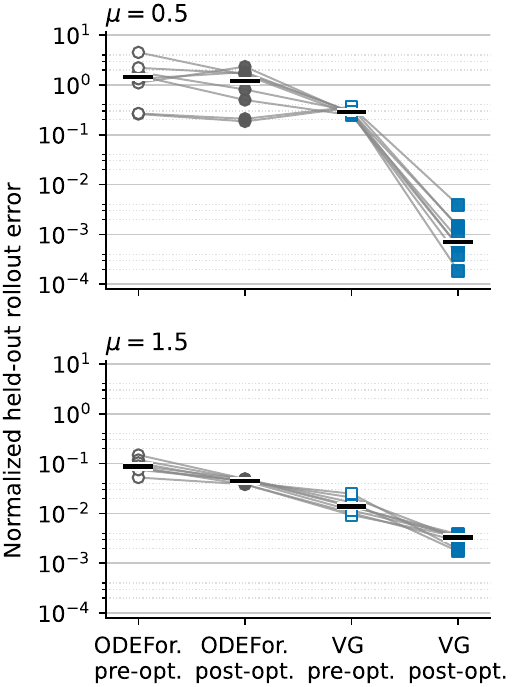}
    \caption{Held-out rollout errors across eight test initial conditions. Circles denote the original ODEFormer protocol and squares the VG workflow; open and filled markers indicate pre- and post-optimization results. Gray lines pair identical initial conditions, and black bars show medians.}
    \label{fig:vdp-errors}
\end{figure}

\paragraph{Results.} Figure~\ref{fig:vdp-errors} shows that the VG model outperforms the single-trajectory protocol for all eight held-out initial conditions at both values of \(\mu\). For \(\mu=0.5\), the median post-optimization error decreases from \(1.21\) to \(6.99\times10^{-4}\); for \(\mu=1.5\), it decreases from \(4.5\times10^{-2}\) to \(3.27\times10^{-3}\). The larger separation at \(\mu=0.5\) is consistent with its longer transient: a single observed orbit constrains less of the surrounding vector field, whereas the VG workflow evaluates a common equation across multiple approaches to the limit cycle. One repeat of the original ODEFormer protocol failed on the test set, and coefficient optimization degraded another.

At each \(\mu\), the candidate with the lowest multi-trajectory rollout error was also the only member of the ten-candidate shortlist to satisfy all verifier checks. Verification therefore in this canonical case distinguished a generalizable closed dynamical model from alternatives that reproduced only individual features, such as the oscillation period or limit-cycle geometry.

After coefficient optimization, Eqs.~\eqref{eq:vdp-mu05}--\eqref{eq:vdp-mu15} were selected:

\begin{equation}
\begin{aligned}
\mu=0.5:\qquad
\dot{x}_0 &= 0.997x_1-0.012x_0,\\
\dot{x}_1 &= 0.542x_1-0.997x_0\\
           &\quad -0.543x_0^2x_1-0.0039x_0x_1,
\end{aligned}
\label{eq:vdp-mu05}
\end{equation}

and

\begin{equation}
\begin{aligned}
\mu=1.5:\quad
\dot{x}_0
  &= 0.996x_1\\
  &\quad +0.000676(0.998+1.273x_0)^2,\\
\dot{x}_1
  &= 1.440x_1-1.002x_0\\
  &\quad -1.461x_0^2x_1-0.0080x_0x_1.
\end{aligned}
\label{eq:vdp-mu15}
\end{equation}

Both systems recover the defining Van der Pol structure: the kinematic relation \(\dot{x}_0\approx x_1\), linear restoring dynamics through \(-x_0\), and the opposing linear and cubic damping terms in \(\dot{x}_1\). Their principal coefficients closely approximate the corresponding ground-truth values, while the additional \(x_0\), \(x_0x_1\), and weak quadratic contributions provide small corrections. The discovered equations are therefore dynamically close approximations rather than exact symbolic replicas of the ground-truth system, supporting evaluation through behavior across initial conditions and physical admissibility rather than expression matching alone.

\subsection{High-Dimensional Physical System Case: Flow Past a Stationary Cylinder at a Fixed Reynolds Number}
\label{sec:fixed-re-results}

The flow-past-a-cylinder case provides the first test of whether the VG workflow can recover compact dynamics from high-dimensional physical fields rather than directly observed ODE states. From a physics standpoint, above the onset of wake instability in this test case configuration, vortices detach alternately from the separated shear layers and form a periodic von K\'arm\'an street. For two-dimensional incompressible flow, the nondimensional spanwise vorticity \(\omega\) satisfies

\begin{equation}
\begin{aligned}
\frac{\partial \omega}{\partial t}
+(\mathbf{u}\cdot\nabla)\omega
&=\frac{1}{Re}\nabla^2\omega,\\
\nabla\cdot\mathbf{u}
&=0,
\qquad
Re=\frac{U_\infty D}{\nu}.
\end{aligned}
\label{eq:vorticity}
\end{equation}

In Eq.~\eqref{eq:vorticity}, $(\mathbf{u}\cdot\nabla)\omega$ represents advection of vorticity by the local flow, whereas $Re^{-1}\nabla^2\omega$ represents viscous diffusion of vorticity gradients. Increasing \(Re\) reduces the relative contribution of this diffusion and permits the wake instability to develop. At \(Re=300\), the fixed value considered here, the two-dimensional simulation exhibits established periodic shedding. As the physical state is a spatial field containing 60,000 vorticity values at each time, dimensionality reduction is required before applying the VG workflow for symbolic model discovery. This is performed using proper orthogonal decomposition (POD) \citep{sirovich1987}.

\paragraph{Experimental setup.}
Vorticity snapshots were generated with the open-source
\texttt{ViscousFlow.jl} solver \citep{eldredgeViscousFlow}, which implements the immersed-layer method \citep{eldredge2022}. A unit-diameter cylinder was centered in \(x/D\in[-1,5]\) and \(y/D\in[-2,2]\), with \(U_\infty=1\), a \(2^\circ\) free-stream incidence, and a no-slip surface. The outputs were interpolated onto a \(300\times200\) grid and sampled at \(\Delta t^*=\Delta tU_\infty/D=0.1\), yielding 1,000 snapshots. Further numerical and preprocessing details are reported by \citet{faraji2025}.

The first 500 snapshots were discarded to isolate established shedding. The remainder was divided chronologically into 250 development, 100 validation, and 150 final-test snapshots. The development interval defined the reduced coordinates and supported candidate generation, ranking, verification, and coefficient optimization; the validation set was used to select the POD representation and decoding configuration before testing on the remaining 150 snapshots.

After subtracting the development-interval mean, snapshot POD produced near-equal mode pairs representing quadrature components of the periodic wake. The first three pairs yielded six standardized coordinates and retained 94.52\% of the fluctuation energy. Validation selected this rank-6 representation and eight decoding windows from the tested ranks \(4\) and \(6\) and window counts \(8\) and \(24\). Candidates decoded from eight five-time-unit windows were pooled and ranked over 24 development windows; the ten highest-ranked were verified, and the selected structure was optimized over the same 24 windows. Technical Appendix~\ref{sec:supp-fixed-re} reports the selection matrix and POD spectrum.

\begin{figure}[b]
    \centering
    \includegraphics[width=0.9\columnwidth]{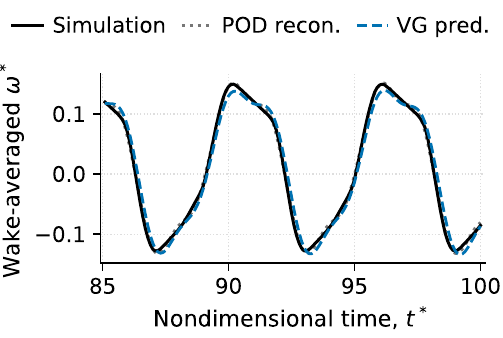}
   \caption{Wake-averaged nondimensional vorticity over the final-test interval at \(Re=300\), evaluated over \(0.5\leq x/D\leq5\) and \(\lvert y/D\rvert\leq1.5\).}
    \label{fig:fixed-re-wake-average}
\end{figure}

\paragraph{Results.} The selected VG model integrated successfully from all eight predeclared starting points in the final-test interval. The window and full-interval coordinate errors in Table~\ref{tab:fixed-re-performance} show that the discovered symbolic system retains the recurrent shedding dynamics across different phases of the cycle. At the field level, the six-mode representation incurs 12.53\% error before symbolic forecasting; the 15.56\% end-to-end result therefore reflects the combined effects of coordinate reduction and dynamics prediction. Figure~\ref{fig:fixed-re-wake-average} shows that the wake-averaged vorticity preserves the dominant phase and amplitude relative to both the simulation and POD reconstruction. Representative vorticity-field snapshot comparisons separating POD truncation from symbolic model forecasting are provided in Technical Appendix~\ref{sec:supp-fixed-re}.

\begin{table}[t]
    \centering
    \setlength{\tabcolsep}{3pt}
    \begin{tabular}{@{}p{0.73\columnwidth}r@{}}
        \toprule
        Quantity & Error (\%)\\
        \midrule
        Median five-time-unit (single-window) coordinate rollout & 3.49\\
        Full-interval coordinate rollout & 6.89\\
        POD field reconstruction & 12.53\\
        VG field relative to POD reconstruction & 8.77\\
        End-to-end field prediction & 15.56\\
        \bottomrule
    \end{tabular}
    \caption{Fixed-\(Re\) final-test performance of the selected VG model. Coordinate rollout errors are mean-square errors normalized by the variance of each standardized POD coordinate. Field errors are temporal means of snapshot-wise relative \(L_2\) errors.}
    \label{tab:fixed-re-performance}
\end{table}

Equation~\eqref{eq:fixed-re-system} gives the coefficient-optimized VG model in standardized rank-6 POD coordinates $z_1,\ldots,z_6$:

\begin{equation}
\begin{aligned}
\dot{z}_1
 &=1.001z_2\\
 &\quad-0.0679(11.746-0.8668z_2)^{-1.052},\\
\dot{z}_2&=-1.123z_1,\\
\dot{z}_3&=-2.099z_4,\\
\dot{z}_4&= 2.120z_3,\\
\dot{z}_5
 &=3.121z_6\\
 &\quad-0.1164\sin(0.1144+11.036z_2),\\
\dot{z}_6&=-0.1458-3.233z_5.
\end{aligned}
\label{eq:fixed-re-system}
\end{equation}

The dominant linear parts of the \((z_1,z_2)\), \((z_3,z_4)\), and \((z_5,z_6)\) subsystems define oscillators with angular frequencies 1.060, 2.109, and 3.177, respectively. Their ratio of \(1:1.99:3.00\) identifies the fundamental shedding cycle and its second and third harmonics. The reciprocal contribution to \(\dot{z}_1\) deforms the fundamental oscillator from exact linear motion, while the sinusoidal dependence of \(\dot{z}_5\) on \(z_2\) couples the fundamental and third-harmonic coordinate pairs. The constant term in \(\dot{z}_6\) produces a small shift in the equilibrium of the third oscillator. Because the POD coordinates are global projections of the vorticity field, these terms describe coupling within the reduced representation and cannot be assigned independently to localized flow mechanisms.

Notably, the VG workflow obtained this system without a wake-specific candidate library or prescribed Navier--Stokes structure: the pretrained backbone searched its learned vocabulary, and verification selected an admissible model reproducing the shedding dynamics.

\subsection{Cross-Parameter Extension: Flow Past a Cylinder across Reynolds Numbers}
\label{sec:multi-re-results}

The fixed-\(Re\) result presented in the previous subsection motivates a more demanding test: whether the VG workflow can derive a single symbolic reduced-order model capable of representing vortex-shedding dynamics as the governing flow parameter, the Reynolds number, varies. 

\paragraph{Experimental setup.}
Model discovery used \(Re\in\{150,200,250,300,350,400,450\}\). The withheld test values were \(Re\in\{175,275,425\}\) for interpolation and \(Re=500\) for extrapolation.

The simulations and field preprocessing followed the fixed-\(Re\) case. After mean subtraction, a common shallow autoencoder (AE) with one 256-neuron hidden layer compressed each 60,000-component snapshot into three latent coordinates. A Reynolds-number coordinate, centered and scaled over the discovery range, was appended and constrained to remain constant during rollout. Candidates generated independently from the seven full-length discovery trajectories were pooled and ranked across all seven; the shortlist was verified, and the selected structure was coefficient-optimized over the same trajectories.

\paragraph{Results.} The selected shallow three-coordinate autoencoder had a mean field-reconstruction error of 5.83\%, the lowest latent roughness (0.0224), and the highest spectral concentration (0.926) among several tested encodings. Roughness measures temporal curvature relative to first-order variation, while spectral concentration measures the fraction of nonzero-frequency power in the three dominant Fourier components. Although several deeper or four-coordinate encodings reconstructed more accurately, among the controlled encodings advanced to VG, only a deeper three-coordinate case yielded a raw admissible equation, with rollout loss 2.17 compared with 1.04 for the selected shallow encoding; coefficient refinement reduced the latter to 0.200 while preserving admissibility. Reconstruction fidelity therefore did not determine symbolic discoverability. Technical Appendix~\ref{sec:supp-ae-sensitivity} reports more details on the outcomes of architecture and initialization-seed sensitivity, latent diagnostics, and corresponding symbolic-model discovery.

In standardized latent coordinates $z_1,z_2,z_3$, with $r$ the normalized Reynolds-number coordinate, Eq.~\eqref{eq:multi-re-system} gives the coefficient-optimized system that satisfied all verifier checks:

\begin{equation}
\begin{aligned}
\dot{z}_1 &= 1.143z_3+0.0422r+0.201z_2z_3,\\
\dot{z}_2 &= 0.798z_1,\\
\dot{z}_3 &=-0.995z_1,\\
\dot{r}   &=0.
\end{aligned}
\label{eq:multi-re-system}
\end{equation}

The \(1.143z_3\) and \(-0.995z_1\) terms define the leading linear oscillator in the \((z_1,z_3)\) coordinates, whose linearized angular frequency is \(\sqrt{1.143\times0.995}=1.066\). The evolution of \(z_2\) is coupled to the same cycle, while the quadratic contribution \(0.201z_2z_3\) feeds this coordinate back into \(\dot{z}_1\), introducing a nonlinear deformation of the underlying oscillator. The additive \(0.0422r\) term shifts the latent vector field with Reynolds number, and \(\dot{r}=0\) preserves the selected parameter value throughout each rollout.

Because \(\dot{z}_3=-(0.995/0.798)\dot{z}_2\), the combination \(z_3+1.247z_2\) is conserved to the precision of the reported coefficients. Consequently, for each fixed value of \(r\), trajectories of the discovered model remain on a two-dimensional invariant surface within the three-dimensional latent space. The Reynolds-number coordinate therefore conditions a shared nonlinear oscillator rather than acting as an independently evolving input.

\begin{table}[t]
    \centering
    {
    \small
    \setlength{\tabcolsep}{2.3pt}
    \begin{tabular}{@{}ccccc@{}}
        \toprule
        \(Re\) &
        \begin{tabular}[c]{@{}c@{}}Latent\\NMSE\end{tabular} &
        \begin{tabular}[c]{@{}c@{}}AE field\\(\%)\end{tabular} &
        \begin{tabular}[c]{@{}c@{}}VG-to-AE\\field (\%)\end{tabular} &
        \begin{tabular}[c]{@{}c@{}}End-to-end\\(\%)\end{tabular}\\
        \midrule
        175 (I) & 0.142 & 7.0 & 19.8 & 20.9\\
        275 (I) & 0.247 & 5.9 & 32.1 & 33.0\\
        425 (I) & 0.043 & 6.0 & 18.9 & 20.7\\
        500 (E) & 0.036 & 6.4 & 11.9 & 13.8\\
        \bottomrule
    \end{tabular}
    }
    \caption{Performance at Reynolds numbers withheld from symbolic discovery. Latent NMSE is the coordinate-wise mean rollout error normalized by the variance of each reference latent coordinate. Field quantities are temporal means of snapshot-wise relative spatial \(L_2\) errors. I and E denote interpolation and extrapolation, respectively.}
    \label{tab:multi-re-performance}
\end{table}

\begin{figure}[t]
    \centering
    \includegraphics[width=\columnwidth]{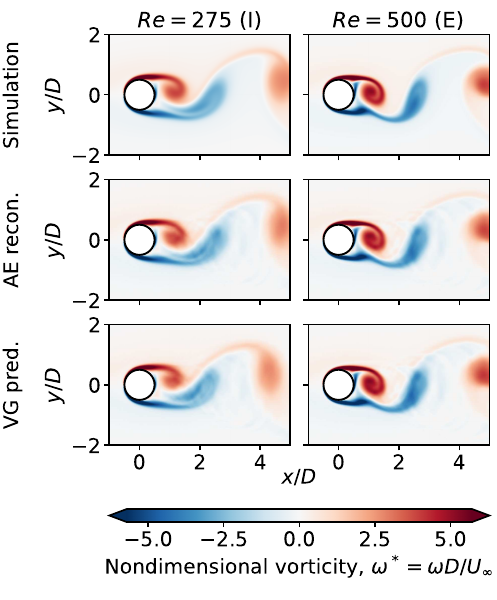}
    \caption{Simulation, autoencoder (AE) reconstruction, and VG-predicted vorticity at \(t^*=75\) for withheld \(Re=275\) (interpolation, I) and \(Re=500\) (extrapolation, E). All panels share a symmetric color scale.}
    \label{fig:multi-re-fields}
\end{figure}

The selected VG model produced finite full-length rollouts at all four withheld Reynolds numbers. Referring to Table~\ref{tab:multi-re-performance}, generalization did not deteriorate monotonically from interpolation to extrapolation: the \(Re=500\) extrapolation case produced the lowest latent NMSE (0.036) and end-to-end field error (13.8\%), whereas the \(Re=275\) interpolation case was the most difficult, with corresponding errors of 0.247 and 33.0\%. Figure~\ref{fig:multi-re-fields} provides important physical context for the latter value. Although displacement of the predicted vortices increases the spatial \(L_2\) error, the VG field retains the alternating wake structure and principal vortex locations. The error is therefore strongly influenced by phase and spatial displacement accumulated during rollout rather than by a loss of the vortex-shedding regime. At \(Re=500\), the close agreement between the autoencoder reconstruction and VG prediction is consistent with its comparatively low VG-to-AE field error of 11.9\%.

Complete latent-coordinate rollouts, wake-averaged vorticity histories, additional field comparisons, and extended error metrics for all four test Reynolds numbers are provided in Technical Appendix~\ref{sec:supp-multi-re}.

\section{Conclusions}

The principal takeaway is that a frozen symbolic transformer pretrained on synthetic ODEs can support physical model discovery beyond its pretraining distribution when paired with multi-trajectory executable verification. Across both the Van der Pol and cylinder-flow cases, transfer depended not only on generating plausible equations but on selecting a common model that satisfied the prescribed dynamical and physical-admissibility tests. Model interpretation is, however, coordinate-dependent: the recovered Van der Pol equations closely approximate the governing structure in directly observed states, whereas the equations discovered in POD and autoencoder coordinates are effective reduced-order laws rather than unique reductions of the Navier--Stokes equations.

The representation results also clarify a central limitation. Symbolic discovery requires the supplied coordinates to admit an approximately closed autonomous description compatible with the backbone's learned vocabulary and pretraining distribution; unresolved modes, memory, forcing, or parameter dependence can prevent this. Pooling and verification enable transfer with a frozen pretrained backbone but cannot supply dynamics or structures absent from its candidate class. These limits motivate tighter neuro-symbolic integration of learned representations and symbolic model discovery \citep{marra2024,cranmer2020}, specifically through joint verified coordinate--dynamics learning that directly integrates executable constraints encoding fundamental physical principles (e.g., conservation, invariance, and symmetry) into search or generation. These directions are particularly important for higher-dimensional, multiscale systems whose reduced coordinates are only approximately closed, including many plasma systems, and would strengthen the basis for interpretable and auditable AI across the natural sciences.

\section*{Code and Data Availability}

The code and data bundle supporting the results reported in this work will be made publicly available upon acceptance for publication.

\FloatBarrier
\bibliographystyle{plainnat}
\bibliography{references}


\clearpage
\onecolumn
\appendix

\renewcommand{\thesection}{S\arabic{section}}
\renewcommand{\thesubsection}{\thesection.\arabic{subsection}}
\renewcommand{\thefigure}{S\arabic{figure}}
\renewcommand{\thetable}{S\arabic{table}}
\renewcommand{\theequation}{S\arabic{equation}}

\renewcommand{\theHsection}{supp.\arabic{section}}
\renewcommand{\theHsubsection}{supp.\arabic{section}.\arabic{subsection}}
\renewcommand{\theHfigure}{supp.\arabic{figure}}
\renewcommand{\theHtable}{supp.\arabic{table}}
\renewcommand{\theHequation}{supp.\arabic{equation}}

\setcounter{figure}{0}
\setcounter{table}{0}
\setcounter{equation}{0}

\section*{Technical Appendices}

These technical appendices provide the supplementary results and analyses referenced in the paper. They report beam-size--sampling-temperature candidate-generation sensitivity
(Section~\ref{sec:supp-beam-temperature}), detailed verifier definitions and their identifiability interpretation (Section~\ref{sec:supp-verifiers}), Van der Pol test-case details (Section~\ref{sec:supp-vdp}), the fixed-Reynolds-number representation study (Section~\ref{sec:supp-fixed-re}), the autoencoder sensitivity analysis (Section~\ref{sec:supp-ae-sensitivity}), and extended results from the multi-Reynolds-number test case (Section~\ref{sec:supp-multi-re}).

\section{Sensitivity to Beam Size and Sampling Temperature}
\label{sec:supp-beam-temperature}

The beam size and sampling temperature are candidate-symbolic-equation generation (decoding) settings inherited from
the symbolic backbone. The beam size determines how many candidate sequences
are retained from each input window, while the temperature controls the
concentration of the sampled token distribution and hence the diversity of
the resulting equations. Because both settings affect the composition and
size of the candidate pool available to VG, their sensitivity was assessed
before the final experiments. The Van der Pol oscillator provides a suitable
diagnostic because its states are observed directly and its governing
structure is known, avoiding the additional influence of coordinate
reduction. This section concerns selection of the decoding settings only;
Section~\ref{sec:supp-vdp} reports the complete Van der Pol trajectory sets,
selected equations, and final-test behavior.

\subsection{Comparison protocol}

Beam sizes 20 and 50 were crossed with temperatures 0.1 and 0.2, giving four
pre-specified decoding configurations. For each of the two oscillator
parameters, $\mu=0.5$ and $1.5$, the same 12 model-construction trajectories
were sampled over $0\leq t\leq20$. Each trajectory contributed an early
window over $0\leq t\leq10$ and a late window over $10\leq t\leq20$, giving
the 24-window model-construction bank. Eight preset members of this bank,
termed \emph{candidate-generation windows} below, were processed individually
by VG to generate equation candidates. These candidates were pooled and
deduplicated before rollout ranking across all 24 model-construction windows
and evaluation by the verifier suite. The same 24-window bank was used for
coefficient optimization of the selected system. Performance was measured on
eight additional validation trajectories that were disjoint from both the
model-construction trajectories and the final-test trajectories. Their exact
initial conditions are reported in Section~\ref{sec:supp-vdp}.

For each candidate-generation window, VG retained a number of sequences equal
to the beam size. The eight candidate-generation windows therefore produced
at most 160 candidates for beam size 20 and 400 for beam size 50 before
canonical deduplication; the resulting pools contained 160 and 398--400
distinct systems, respectively. Each pool was ranked by its mean rollout
error across the 24 model-construction windows. Section~\ref{sec:supp-verifiers}
defines the verifier metrics and acceptance decisions, and
Table~\ref{tab:supp-verifier-settings} summarizes their numerical settings for
the Van der Pol and cylinder-flow cases. The ten highest-ranked systems were
evaluated using these verifiers. The admissible system with the lowest rollout
error was coefficient-optimized over the same 24-window bank and then
verified again. Random seeds, trajectories, window selections, verifier
settings, and optimization procedure were identical across the four
configurations. When no member of a shortlist passed the pre-optimization
verifiers, that outcome was retained; the rollout-ranked candidate was
optimized only to complete the diagnostic comparison.

\subsection{Validation performance and admissibility}

Table~\ref{tab:supp-beam-temperature} reports the validation error of the
lowest-ranked pre-optimization candidate satisfying the verifier suite and
the corresponding post-optimization result. The error is the coordinate-wise
mean rollout error normalized by the variance of each reference state. The
configuration with beam size 20 and temperature 0.1 produced an admissible
system before and after coefficient optimization at both values of $\mu$, with
the lowest post-optimization validation median in each case. Increasing the
temperature at beam size 20 increased the remaining error, particularly for
$\mu=1.5$.
The larger beam did not provide a consistent benefit: at temperature 0.1 no
pre-optimization candidate was admissible for $\mu=0.5$, while the
post-optimization system for $\mu=1.5$ no longer satisfied the local
vector-field verifier. Beam size 50 also expanded the candidate pool by
approximately a factor of 2.5 and correspondingly increased the rollout
ranking burden.

\begin{table}[!ht]
  \centering
  \caption{Sensitivity of validation rollout error and verifier
  admissibility to beam size and sampling temperature. ``Pre-opt. candidate''
  indicates whether at least one member of the ten-system shortlist passed
  every applicable verifier. Bold rows identify the configuration carried
  forward.}
  \label{tab:supp-beam-temperature}
  \small
  \setlength{\tabcolsep}{4.0pt}
  \begin{tabular}{ccccccc}
    \toprule
    Beam & Temperature & $\mu$ & Pre-opt. candidate &
    Pre-opt. NMSE & Post-opt. NMSE & Post-opt. admissible \\
    \midrule
    \textbf{20} & \textbf{0.1} & \textbf{0.5} & \textbf{Yes} &
    \textbf{0.288} & $\mathbf{7.97\times10^{-4}}$ & \textbf{Yes} \\
    \textbf{20} & \textbf{0.1} & \textbf{1.5} & \textbf{Yes} &
    \textbf{0.0145} & $\mathbf{2.74\times10^{-3}}$ & \textbf{Yes} \\
    20 & 0.2 & 0.5 & Yes & 0.224 & $1.60\times10^{-3}$ & Yes \\
    20 & 0.2 & 1.5 & Yes & 0.756 & 0.0255 & Yes \\
    50 & 0.1 & 0.5 & No$^{*}$ & 0.155$^{*}$ & 0.0320$^{*}$ & Yes$^{*}$ \\
    50 & 0.1 & 1.5 & Yes & 0.331 & 0.0249 & No \\
    50 & 0.2 & 0.5 & Yes & 0.136 & $8.81\times10^{-4}$ & Yes \\
    50 & 0.2 & 1.5 & Yes & 0.366 & 0.0205 & Yes \\
    \bottomrule
  \end{tabular}
  \parbox{\linewidth}{\footnotesize\raggedright
  $^{*}$No shortlist member passed all pre-optimization verifiers. The
  reported errors and post-optimization decision describe the
  rollout-ranked candidate retained for diagnostic purposes; this candidate
  was not treated as a verifier-guided pre-optimization selection.}
\end{table}
\FloatBarrier

These results support beam size 20 and temperature 0.1 as a reasonable
operating point rather than indicating that the decoding settings are
immaterial. This configuration provided the most accurate admissible
post-optimization system for both oscillator parameters while using the
smaller candidate pool. It was consequently fixed before the final-test comparison reported in Section~\ref{sec:vdp-results} and was carried forward to the cylinder-flow experiments.

\subsection{Early- and late-window diagnostic}

The selected decoding configuration was also examined separately on the 12
early and 12 late model-construction windows to determine whether its
aggregate rollout score concealed a marked difference between transient-rich
and established-cycle behavior. This comparison did not introduce an
additional selection rule. For $\mu=0.5$, the median pre-optimization NMSE was
0.114 on the early windows and 0.0982 on the late windows; coefficient
optimization reduced these values to $7.39\times10^{-4}$ and
$3.19\times10^{-4}$, respectively. For $\mu=1.5$, the corresponding medians
were 0.00537 and 0.00266 before optimization and $1.26\times10^{-3}$ and
$8.07\times10^{-4}$ afterwards. The higher errors on the early windows are
consistent with their inclusion of trajectories approaching the limit cycle,
which sample a broader portion of the surrounding vector field. The low
post-optimization errors in both regimes show that the selected configuration
was supported by accurate rollouts across both the transient-rich early
windows and the more regularly oscillatory late windows.

\section{Verifier Definitions, Identifiability, and Sensitivity Considerations}
\label{sec:supp-verifiers}

\subsection{Constrained model selection}

The verifier-guided workflow described in Section~\ref{sec:method} of the paper separates dynamical admissibility from rollout-based ranking. Let
$\mathcal{C}_{K}$ denote the shortlist obtained after pooling,
deduplicating, and ranking candidates across the discovery trajectories.
For a candidate $\mathbf{h}\in\mathcal{C}_{K}$, verifier $\ell$ evaluates a
metric $m_{\ell}(\mathbf{h})$ against a tolerance $\tau_{\ell}$. An
inequality-based decision can be written compactly as
\begin{equation}
  \mathsf{pass}_{\ell}(\mathbf{h})
  =
  \mathbb{I}\!\left[
    m_{\ell}(\mathbf{h})\leq\tau_{\ell}
  \right],
  \label{eq:binary-verifier}
\end{equation}
where $\mathbb{I}[\cdot]$ is the indicator function and $\tau_{\ell}$ is the
largest accepted value of the metric. This form applies because the numerical
metrics used here measure discrepancy, variability, drift, or growth, for
which smaller values indicate closer agreement or greater admissibility.
Conditions on symbolic structure, such as an identically zero parameter
equation, are evaluated directly. The admissible set and selected model are
then
\begin{equation}
  \mathcal{A}
  =
  \left\{
    \mathbf{h}\in\mathcal{C}_{K}:
    \mathsf{pass}_{\ell}(\mathbf{h})=1
    \ \text{for every applicable verifier }\ell
  \right\},
  \qquad
  \mathbf{h}^{\star}
  =
  \underset{\mathbf{h}\in\mathcal{A}}{\arg\min}\,
  \mathcal{L}_{\mathrm{roll}}(\mathbf{h}).
  \label{eq:constrained-selection}
\end{equation}
Equations~\eqref{eq:binary-verifier} and
\eqref{eq:constrained-selection} therefore define admissibility separately
from $\mathcal{L}_{\mathrm{roll}}$. Rollout error orders only the candidates
that satisfy every applicable requirement. When $\mathcal{A}$ is empty, no
admissible symbolic model is selected.

\subsection{Verifier definitions and interpretation}

\paragraph{Local vector-field agreement.}
Let $\boldsymbol{\xi}_{i}^{(m)}\in\mathbb{R}^{d_{\xi}}$ be the full state at
sample $i$ of discovery trajectory $m$. Its numerically estimated rate in
evaluated coordinate $j=1,\ldots,d$ is the scalar
$\dot{\xi}_{ij}^{(m)}$, while
$h_j(\boldsymbol{\xi}_{i}^{(m)})$ is the corresponding scalar component of
the candidate vector field evaluated at the full state. The local discrepancy
and its admissibility condition are given by
\begin{equation}
  \begin{aligned}
  V_{\mathrm{loc}}(\mathbf{h})
  &=
  \frac{1}{d}\sum_{j=1}^{d}
  \frac{
  \operatorname{MSE}_{m,i}
  \left(h_j(\boldsymbol{\xi}_{i}^{(m)}),
  \dot{\xi}_{ij}^{(m)}\right)
  }{
  \operatorname{Var}_{m,i}
  \left(\dot{\xi}_{ij}^{(m)}\right)+\epsilon
  },\\
  \mathsf{pass}_{\mathrm{loc}}(\mathbf{h})
  &=
  \mathbb{I}\!\left[
  V_{\mathrm{loc}}(\mathbf{h})
  \leq
  2\min_{\widetilde{\mathbf{h}}\in\mathcal{C}_{K}}
  V_{\mathrm{loc}}(\widetilde{\mathbf{h}})
  \right],
  \end{aligned}
  \label{eq:local-verifier}
\end{equation}
where $\epsilon=10^{-12}$ prevents division by zero. For the directly
observed and fixed-parameter systems, $d_{\xi}=d$; for the cross-$\Rey$
system, $d_{\xi}=d+1$ because the input also contains the normalized Reynolds
coordinate $r=(\Rey-300)/150$, while the score is evaluated over the $d$
latent-state equations. The variance normalization in
Eq.~\eqref{eq:local-verifier} gives each evaluated coordinate comparable
influence despite different derivative scales. Because derivative scale and numerical
differentiation error also differ between the observed Van der Pol states,
POD coordinates, and autoencoder coordinates, the tolerance is defined
relative to the lowest local discrepancy within the same shortlist. A
candidate passes when its instantaneous direction and rate remain within a
factor of two of that case-specific reference.

\paragraph{Long-horizon boundedness.}
Let $\widehat{\boldsymbol{\xi}}_{\mathbf h}^{(s)}(t)$ denote the trajectory
obtained by integrating candidate $\mathbf h$ from observed starting state
$s$, and let
$B_{\mathrm{obs}}=\max_{m,i}\|\boldsymbol{\xi}_{i}^{(m)}\|_{2}$ denote the
largest state norm in the corresponding reference data. Boundedness requires
\begin{equation}
  \mathsf{pass}_{\mathrm{bound}}(\mathbf{h})
  =
  \mathbb{I}\!\left[
    \widehat{\boldsymbol{\xi}}_{\mathbf h}^{(s)}(t)
    \text{ is finite for all }s,t
    \ \land\
    \max_{s,t}
    \left\|
      \widehat{\boldsymbol{\xi}}_{\mathbf h}^{(s)}(t)
    \right\|_2
    \leq c_{\mathrm{bound}}B_{\mathrm{obs}}
  \right].
  \label{eq:boundedness-verifier}
\end{equation}
In Eq.~\eqref{eq:boundedness-verifier}, the norm and its reference scale are
evaluated only over the latent coordinates and separately at each Reynolds
number for the cross-$\Rey$ case.
The factor is $c_{\mathrm{bound}}=1.5$ for Van der Pol and
$c_{\mathrm{bound}}=3$ for the reduced flow coordinates. This check rejects
numerically divergent solutions and trajectories whose state magnitude grows
well beyond that represented by the observations.

\paragraph{Oscillation amplitude.}
After removal of the initial verification interval, the amplitude of
coordinate $j$ is estimated by the robust percentile half-range
$A_j(\boldsymbol{\xi})=
\tfrac{1}{2}[Q_{0.99}(\xi_j)-Q_{0.01}(\xi_j)]$.
Let $A_j^{(m)}$ be the amplitude of reference trajectory $m$, and let
$\widehat{A}_{s,j}^{(m)}$ be the corresponding amplitude of the candidate
rollout from starting state $s$. If $\mathcal{M}_{v}$ indexes the reference
cases used for verification and $\mathcal{J}_{A}$ the coordinates whose
amplitudes are evaluated, then
\begin{equation}
  \begin{aligned}
  m_{\mathrm{amp}}(\mathbf h)
  &=
  \frac{1}{|\mathcal{M}_{v}|\,|\mathcal{J}_{A}|}
  \sum_{m\in\mathcal{M}_{v}}
  \sum_{j\in\mathcal{J}_{A}}
  \frac{
    \left|
      \operatorname{median}_{s}\widehat{A}_{s,j}^{(m)}
      -A_j^{(m)}
    \right|
  }{|A_j^{(m)}|+\epsilon},\\
  \mathsf{pass}_{\mathrm{amp}}(\mathbf{h})
  &=
  \mathbb{I}\!\left[m_{\mathrm{amp}}(\mathbf h)\leq0.15\right].
  \end{aligned}
  \label{eq:amplitude-verifier}
\end{equation}
For Van der Pol, $\mathcal{J}_{A}$ contains the observed oscillatory
coordinate, and $\mathcal{M}_{v}$ contains one pooled reference case whose
target is formed from the discovery trajectories. For the fixed-$\Rey$ flow
case, $\mathcal{M}_{v}$ contains the single $\Rey=300$ reference trajectory;
for the cross-$\Rey$ case, it contains the three reference trajectories at
$\Rey=150,300,$ and $450$. All retained coordinates belong to
$\mathcal{J}_{A}$ in both flow cases. The verifier therefore tests whether
the candidate reproduces the size of the recurrent motion, not only its
pointwise path, as expressed by Eq.~\eqref{eq:amplitude-verifier}.

\paragraph{Period or dominant frequency.}
Let $q_j^{(m)}$ denote the characteristic temporal scale of coordinate $j$ in
reference case $m$, and let $\widehat q_{s,j}^{(m)}$ denote the same quantity
from the candidate rollout. Its relative discrepancy is
\begin{equation}
  \begin{aligned}
  m_{\mathrm{time}}(\mathbf h)
  &=
  \frac{1}{|\mathcal{M}_{v}|\,|\mathcal{J}_{q}|}
  \sum_{m\in\mathcal{M}_{v}}
  \sum_{j\in\mathcal{J}_{q}}
  \frac{
    \left|
      \operatorname{median}_{s}\widehat q_{s,j}^{(m)}
      -q_j^{(m)}
    \right|
  }{|q_j^{(m)}|+\epsilon},\\
  \mathsf{pass}_{\mathrm{time}}(\mathbf{h})
  &=
  \mathbb{I}\!\left[m_{\mathrm{time}}(\mathbf h)\leq0.10\right].
  \end{aligned}
  \label{eq:frequency-verifier}
\end{equation}
For Van der Pol, $q$ is the median interval between successive peaks after
the transient and $\mathcal{J}_{q}$ contains the observed oscillatory
coordinate. For each reduced flow coordinate, $q_j$ is instead the dominant
nonzero Fourier frequency,
$q_j=\operatorname*{arg\,max}_{\nu>0}
|\mathcal{F}[\xi_j-\overline{\xi}_j](\nu)|^2$.
Peak spacing measures the period of the directly observed oscillator,
whereas the spectral definition treats all reduced flow coordinates
uniformly. In both cases, Eq.~\eqref{eq:frequency-verifier} tests preservation
of the dominant oscillation rate.

\paragraph{Starting-state consistency.}
Each candidate is integrated from three observed starting states. For a
rollout property $u\in\{A,q\}$, where $A$ is amplitude and $q$ is period or
dominant frequency, define
\begin{equation}
  \begin{aligned}
  C_u(\mathbf h)
  &=
  \frac{1}{|\mathcal{M}_{v}|\,|\mathcal{J}_{u}|}
  \sum_{m\in\mathcal{M}_{v}}
  \sum_{j\in\mathcal{J}_{u}}
  \frac{
    \operatorname{std}_{s}
    \left(\widehat u_{s,j}^{(m)}\right)
  }{
    \operatorname{mean}_{s}
    \left|\widehat u_{s,j}^{(m)}\right|+\epsilon
  },\\
  \mathsf{pass}_{\mathrm{cons}}(\mathbf h)
  &=
  \mathbb{I}\!\left[
    C_A(\mathbf h)\leq0.10
    \ \land\
    C_q(\mathbf h)\leq0.05
  \right].
  \end{aligned}
  \label{eq:consistency-verifier}
\end{equation}
Here the hat denotes a quantity measured from an integrated candidate
trajectory, and the standard deviation and mean are taken across the three
starting states. A stable recurrent model should approach comparable
amplitudes and time scales when initiated at different observed phases.
Equation~\eqref{eq:consistency-verifier} excludes candidates for which those
long-time properties depend strongly on the selected starting state.

\paragraph{Parameter conservation and participation.}
For the cross-$\Rey$ model, let $h_r$ denote the candidate equation for the
normalized Reynolds coordinate defined above and let $h_j$,
$j=1,\ldots,d_z$, denote the equations for the $d_z$ latent coordinates. The
parameter check is
\begin{equation}
  \mathsf{pass}_{\mathrm{par}}(\mathbf{h})
  =
  \mathbb{I}\!\left[
    h_r\equiv0
    \ \land\
    \max_{s,t}|\,\widehat r_{\mathbf h}^{(s)}(t)-r^{(s)}(0)\,|
    \leq10^{-8}
    \ \land\
    \exists\,j\leq d_z:
    \frac{\partial h_j}{\partial r}\not\equiv0
  \right].
  \label{eq:parameter-verifier}
\end{equation}
The first condition preserves the autonomous augmentation
$\dot r=0$ symbolically, and the second confirms the same conservation in
numerical rollouts. The third requires the latent dynamics to depend
explicitly on $r$; carrying an unused constant coordinate is not sufficient
for a shared cross-parameter model.

Table~\ref{tab:supp-verifier-settings} summarizes the settings associated with Eqs.~\eqref{eq:local-verifier}--\eqref{eq:parameter-verifier}.

The cross-$\Rey$ shortlist is the deduplicated union of the ten candidates with
the lowest multi-trajectory rollout errors and the ten lowest-error
candidates containing explicit Reynolds-number dependence in at least one
latent-state equation. This two-branch construction retains hypotheses
capable of representing a shared parametric law when rollout ranking favors a parameter-independent approximation over the discovery trajectories.
Overlap between the branches produced the 17 distinct candidates reported in
Table~\ref{tab:supp-verifier-settings}.

For the cross-$\Rey$ integration, one model-time unit corresponds to
$0.1$ physical-time units. The numerical
horizon and discard interval of 300 and 50 model-time units therefore
correspond to the reported values of 30 and 5.

The numerical thresholds in Table~\ref{tab:supp-verifier-settings} were held
constant within each experimental setting. Their sensitivity is examined for
the representative fixed-$\Rey$ case in
Section~\ref{sec:supp-tolerance-sensitivity}.

\begin{table}[!h]
  \centering
  \caption{Verifier settings used in the three experimental settings.}
  \label{tab:supp-verifier-settings}
  \begin{tabularx}{\linewidth}{lXXX}
    \toprule
    Setting & Van der Pol & Fixed $\Rey=300$ & Cross-$\Rey$ \\
    \midrule
    Shortlist size & 10 & 10 & $17^{*}$ \\
    Local vector-field factor & 2.0 & 2.0 & 2.0 \\
    Bound factor & 1.5 & 3.0 & 3.0 \\
    Verification horizon & 60 & 30 & $30^{**}$ \\
    Discard interval & 30 & 5 & $5^{**}$ \\
    Amplitude tolerance & 0.15 & 0.15 & 0.15 \\
    Period/frequency tolerance & 0.10 & 0.10 & 0.10 \\
    Amplitude CV tolerance & 0.10 & 0.10 & 0.10 \\
    Period/frequency CV tolerance & 0.05 & 0.05 & 0.05 \\
    Parameter-drift tolerance & NA & NA & $10^{-8}$ \\
    \bottomrule
  \end{tabularx}
  \parbox{\linewidth}{\footnotesize\raggedright
  $^{*}$The two cross-$\Rey$ shortlist branches yielded 17 distinct
  candidates.\par
  $^{**}$Cross-$\Rey$ horizon and discard values are reported in
  physical-time units.}
\end{table}
\FloatBarrier

\subsection{Identifiability interpretation}
\label{sec:supp-identifiability}

Identifiability concerns whether the observations and modeling assumptions
distinguish one structure or coefficient vector from plausible alternatives.
This issue is central to trajectory-based symbolic discovery because
different equations can reproduce a finite observed path while defining
different vector fields away from it, different long-time attractors, or
different responses to a changed initial condition or parameter. A low
rollout error on the supplied trajectories is therefore not, by itself,
evidence that the dynamics have been uniquely identified.

The claim made in this work is intentionally restricted. The verifier suite
does not establish global structural identifiability over an unrestricted
class of differential equations. It improves discrimination within the
finite hypothesis class generated by the symbolic backbone and over the
specified state domains, initial conditions, parameter values, and
verification horizons. To express this distinction, let
$\mathcal{H}_{G}$ be the generated, deduplicated hypothesis class,
$R(\mathbf f)$ its multi-trajectory rollout discrepancy, and
$\varepsilon$ a chosen rollout tolerance. The set of candidates that are
indistinguishable at this rollout resolution is
\begin{equation}
  \mathcal{E}_{R}(\varepsilon)
  =
  \left\{
  \mathbf{f}\in\mathcal{H}_{G}:
  R(\mathbf{f})\leq\varepsilon
  \right\}.
  \label{eq:rollout-equivalence}
\end{equation}
Equation~\eqref{eq:rollout-equivalence} describes equivalence only with
respect to the sampled rollout discrepancy.
Let $\mathcal{H}_{A}\subseteq\mathcal{H}_{G}$ denote the candidates satisfying
all applicable dynamical and physical-admissibility conditions. Verification
then restricts the rollout-consistent set to
\begin{equation}
  \mathcal{E}_{VG}(\varepsilon)
  =
  \mathcal{E}_{R}(\varepsilon)\cap\mathcal{H}_{A}.
  \label{eq:verified-equivalence}
\end{equation}
The intersection in Eq.~\eqref{eq:verified-equivalence} improves practical
identifiability when it removes at least one rollout-consistent alternative
while retaining an admissible candidate. The additional discrimination comes
from properties not represented by a scalar rollout ordering alone,
including local vector-field agreement, long-horizon boundedness, recurrent
behavior across starting states, and parameter conservation and
participation.

\subsection{Local sensitivity-rank interpretation}

The preceding set-based view compares distinct candidates in a finite
hypothesis class. A complementary local view asks whether the rollout and
verification diagnostics can distinguish small coefficient changes within
one symbolic structure. Let
$\boldsymbol{\theta}\in\mathbb{R}^{n_{\theta}}$ contain its
$n_{\theta}$ free coefficients,
$\mathbf y(\boldsymbol{\theta})$ collect the sampled candidate rollouts, and
$\mathbf m(\boldsymbol{\theta})$ collect the real-valued verifier metrics
before thresholding. The mapping from coefficients to the combined vector of
rollout samples and verifier metrics has the local sensitivity matrix
\begin{equation}
  \mathbf{J}_{VG}(\boldsymbol{\theta})
  =
  \begin{bmatrix}
    \partial\mathbf{y}/\partial\boldsymbol{\theta}\\
    \partial\mathbf{m}/\partial\boldsymbol{\theta}
  \end{bmatrix}.
  \label{eq:combined-sensitivity}
\end{equation}
Each column of $\mathbf J_{VG}$ in
Eq.~\eqref{eq:combined-sensitivity} describes how the measured quantities
change under a perturbation of one coefficient. Appending verifier
sensitivities cannot reduce the rank available from the rollout samples:
\begin{equation}
  \operatorname{rank}\mathbf{J}_{VG}
  \geq
  \operatorname{rank}
  \left(
  \partial\mathbf{y}/\partial\boldsymbol{\theta}
  \right).
  \label{eq:rank-inequality}
\end{equation}
A strict inequality in Eq.~\eqref{eq:rank-inequality} means that at least one
coefficient direction unresolved by the sampled rollouts affects an
additional diagnostic. Full column rank is a sufficient local condition for
coefficient identifiability within the fixed structure, subject to numerical
conditioning and symbolic symmetries. This rank analysis provides an
interpretation of the information contributed by verification; it was not an
additional computational step in the experiments. The binary decisions
themselves are not differentiated, and metrics based on peak locations,
percentiles, or discrete spectral maxima are interpreted locally only while
the selected features remain unchanged.

\subsection{Implications for joint coordinate and equation discovery}
\label{sec:supp-joint-discovery}

In the present workflow, POD or autoencoder coordinates are selected before
symbolic discovery, so identifiability is assessed conditionally on that
representation. A future joint formulation could instead allow verification
information to influence the coordinates themselves. Let
$\mathcal{E}_{\boldsymbol{\phi}}$ be an encoder with parameters
$\boldsymbol{\phi}$,
$\mathbf z_{\boldsymbol{\phi}}=
\mathcal{E}_{\boldsymbol{\phi}}(\boldsymbol{\omega})$ its latent state, and
$\mathbf g_{\boldsymbol{\theta}}$ a symbolic latent vector field with
coefficients $\boldsymbol{\theta}$. The sensitivity of reconstruction,
latent rollout, and verifier quantities to both parameter sets could be
represented by
\begin{equation}
  \mathbf{J}_{\mathrm{joint}}
  =
  \frac{\partial}{
    \partial(\boldsymbol{\phi},\boldsymbol{\theta})
  }
  \begin{bmatrix}
    \mathcal{R}_{\boldsymbol{\phi}}
      (\mathbf z_{\boldsymbol{\phi}})\\
    \widehat{\mathbf z}_{\boldsymbol{\phi},\boldsymbol{\theta}}(t)\\
    \mathbf m(\boldsymbol{\phi},\boldsymbol{\theta})
  \end{bmatrix},
  \label{eq:joint-coordinate-sensitivity}
\end{equation}
where $\mathcal R_{\boldsymbol{\phi}}$ is the reconstruction map,
$\widehat{\mathbf z}_{\boldsymbol{\phi},\boldsymbol{\theta}}(t)$ is the
integrated latent trajectory, and $\mathbf m$ contains differentiable
surrogates of the verifier properties. Equation~\eqref{eq:joint-coordinate-sensitivity}
shows how additional dynamical information could distinguish embeddings
that reconstruct the field similarly but differ in approximate closure,
boundedness, recurrence, or parameter dependence. Such a formulation could
improve the identifiability of latent coordinates compatible with symbolic
discovery. It remains a prospective extension; the autoencoder used here was
optimized without verifier feedback.

\subsection{Parameter dependence in the
\texorpdfstring{cross-$\Rey$}{cross-Re} model}

In the cross-$\Rey$ experiment, each decoding is performed for one trajectory
whose appended normalized Reynolds coordinate $r$ is constant. A single
trajectory therefore cannot identify how the latent dynamics vary with
Reynolds number. For example, on a trajectory with $r=r_0\neq0$, the terms
\begin{equation}
  \alpha z_j
  =
  \frac{\alpha}{r_0}\,r z_j
  \qquad\text{when }r=r_0
  \label{eq:single-trajectory-parameter-ambiguity}
\end{equation}
are observationally indistinguishable. The occurrence of $r$ in an
individually decoded expression is therefore a candidate structural
hypothesis, rather than evidence that one decoding has resolved the
cross-trajectory parameter dependence.
The ambiguity in Eq.~\eqref{eq:single-trajectory-parameter-ambiguity} is
resolved only by evaluating a shared candidate across different values of
$r$.

Such terms can nevertheless arise because $r$ is supplied as a state
variable and the symbolic pretraining distribution includes cross-variable
products and related dependencies. Variations in latent amplitude,
frequency, and trajectory geometry at different Reynolds numbers can thus
be represented by expressions involving the constant parameter channel.
The cross-trajectory evidence enters after decoding: candidates are pooled
and evaluated as shared equations over all seven discovery Reynolds numbers.
The parameter verifier in Eq.~\eqref{eq:parameter-verifier} additionally
requires conservation of $r$ and its participation in at least one latent
equation. Parameter dependence in the selected model consequently emerges
from candidate generation followed by cross-$\Rey$ pooling, evaluation, and
verification, rather than from simultaneous observation of all trajectories
during an individual decode.

\subsection{Sensitivity to verifier tolerances}
\label{sec:supp-tolerance-sensitivity}

The numerical tolerances in Table~\ref{tab:supp-verifier-settings} determine
the boundary of the admissible set and therefore warrant a sensitivity
check. We use the fixed-$\Rey=300$ cylinder case as a representative example
because it combines the local, boundedness, amplitude, frequency, and
starting-state verifiers without the additional parameter constraints of the
cross-$\Rey$ model. The same perturbation procedure can be applied to the Van
der Pol and cross-$\Rey$ cases, although stability in this representative
case does not imply invariance of their admissible sets.

The sensitivity matrix crosses POD ranks four and six with candidate
generation from either 8 evenly distributed development windows or all 24
development windows. The two even POD ranks retain complete modal pairs while
testing a more compact and a more energetic representation; the two window
counts compare selective and exhaustive candidate generation from the same
window bank. Candidate ranking, verification, and coefficient refinement use
all 24 windows in every configuration. Section~\ref{sec:supp-fixed-re-matrix}
provides the full representation and selection comparison.

Let $\gamma\in\{0.75,1.0,1.5\}$ jointly scale the local vector-field
agreement factor in Eq.~\eqref{eq:local-verifier} and the amplitude,
frequency, amplitude-CV, and frequency-CV tolerances. Thus
$\gamma<1$ imposes stricter quantitative agreement, $\gamma=1$ recovers the
nominal settings in Table~\ref{tab:supp-verifier-settings}, and $\gamma>1$
relaxes them. Finite integration and boundedness remain at their nominal
settings because they serve a different role: finite integration is a
Boolean requirement for a valid rollout, while the boundedness limit defines
the admissible state envelope relative to the observed dynamics. Changing
either would alter the minimum dynamical-admissibility requirement rather
than isolate sensitivity to the agreement tolerances.

The number of admissible candidates, ordered by
$\gamma=(0.75,1.0,1.5)$, was:
\begin{itemize}
  \item \textbf{POD rank four, 8 decoding windows:} $(10,10,10)$;
  \item \textbf{POD rank four, 24 decoding windows:} $(10,10,10)$;
  \item \textbf{POD rank six, 8 decoding windows:} $(5,6,9)$; and
  \item \textbf{POD rank six, 24 decoding windows:} $(9,10,10)$.
\end{itemize}
The admissible set therefore expands under looser tolerances for the
rank-six cases, whereas every shortlisted rank-four candidate satisfies even
the stricter setting. In all four configurations, however, the
lowest-rollout admissible equation selected at the nominal tolerance remained
the selected equation throughout the sensitivity range. The nominally
selected pre-optimization equation and the next-lowest-rollout admissible
alternative for the selected rank-six, eight-window configuration are
reported in Section~\ref{sec:supp-fixed-re-matrix}, together with the complete
selection matrix.

These results show that the verifier thresholds influence which secondary
candidates enter the admissible set, as expected for any thresholded
selection rule, but do not materially change the final outcome in this
representative fixed-$\Rey$ case. The nominal values provide a reasonable
balance: they reject candidates with substantial local or oscillatory
disagreement without requiring near-exact agreement from an effective
reduced-order model. Tightening or relaxing them by 25\% and 50\%,
respectively, preserves the selected equation in every configuration.

\section{Van der Pol Test-Case Details and Extended Results}
\label{sec:supp-vdp}

Section~\ref{sec:supp-beam-temperature} used the Van der Pol oscillator to
select the beam size and sampling temperature inherited from the symbolic
backbone. The present section documents the final experiment conducted after
those settings had been fixed. It provides the exact trajectory sets and
their evaluation roles,
the selected pre- and post-optimization equations, their verifier outcomes,
and representative final-test rollouts. The candidate-generation (decoding)
comparison is not repeated here: the results below use only the selected
beam size of 20 and temperature of 0.1.

\subsection{Trajectory Sets and Evaluation Roles}

For both $\mu=0.5$ and $1.5$, trajectories were integrated over
$0\leq t\leq20$ at 301 equally spaced times. The same initial-condition
sets were used at both parameter values. The complete fixed-seed sets and
their distinct roles are shown below: the model-construction bank is on the
left, the validation set used in the beam--temperature comparison is in the
center, and the final-test set is on the right.

\begin{center}
\small
\begin{minipage}[t]{0.42\linewidth}
\centering
\textbf{Model-construction bank}\\[3pt]
\setlength{\tabcolsep}{3.2pt}
\begin{tabular}{r rr c}
  \toprule
  $j$ & $x_0(0)$ & $x_1(0)$ & Candidate source \\
  \midrule
   0 &  0.023643 &  0.900927 & -- \\
   1 & -0.711681 &  0.897299 & late \\
   2 & -0.376337 & -0.153347 & early, late \\
   3 &  0.655405 & -0.181602 & -- \\
   4 &  0.099187 & -0.944882 & -- \\
   5 &  0.507026 &  0.076287 & late \\
   6 & -0.340537 &  0.576857 & late \\
   7 & -0.393610 & -0.093004 & -- \\
   8 & -0.731917 & -0.193774 & early, late \\
   9 & -0.593090 & -0.475373 & early \\
  10 &  0.500729 & -0.439182 & -- \\
  11 & -0.029618 &  0.961474 & -- \\
  \bottomrule
\end{tabular}
\par\vspace{3pt}
\raggedright\footnotesize
Early window: $0\leq t\leq10$ (151 samples).\\
Late window: $10\leq t\leq20$ (151 samples).
\end{minipage}\hfill
\begin{minipage}[t]{0.27\linewidth}
\centering
\textbf{Beam--temperature validation}\\[3pt]
\setlength{\tabcolsep}{3.2pt}
\begin{tabular}{r rr}
  \toprule
  $j$ & $x_0(0)$ & $x_1(0)$ \\
  \midrule
  0 & -0.476776 & -0.403018 \\
  1 &  0.628451 & -0.816168 \\
  2 &  0.200201 &  0.457121 \\
  3 & -0.624198 & -0.889707 \\
  4 & -0.450061 &  0.314866 \\
  5 &  0.124531 & -0.699875 \\
  6 & -0.134738 &  0.338595 \\
  7 & -0.154431 &  0.266369 \\
  \bottomrule
\end{tabular}
\end{minipage}\hfill
\begin{minipage}[t]{0.27\linewidth}
\centering
\textbf{Final test}\\[3pt]
\setlength{\tabcolsep}{3.2pt}
\begin{tabular}{r rr}
  \toprule
  $j$ & $x_0(0)$ & $x_1(0)$ \\
  \midrule
  0 & -0.828702 & -0.526379 \\
  1 &  0.602549 &  0.164324 \\
  2 & -0.811743 & -0.133746 \\
  3 & -0.041897 & -0.680522 \\
  4 &  0.469154 & -0.772656 \\
  5 & -0.217544 &  0.033480 \\
  6 & -0.138744 &  0.173597 \\
  7 &  0.475676 &  0.912535 \\
  \bottomrule
\end{tabular}
\end{minipage}
\end{center}

As shown in the left-most table, the 12 model-construction trajectories each
contributed one early and one late window, giving a 24-window bank. The eight
windows used as inputs for candidate generation were fixed before the
beam--temperature comparison: trajectory 1 contributed its late window;
trajectory 2 contributed both windows; trajectories 5 and 6 contributed
their late windows; trajectory 8 contributed both windows; and trajectory 9
contributed its early window. The saved bank orders the windows by
trajectory, with each trajectory's early window immediately followed by its
late window. Under this storage convention, the eight candidate-generation
inputs have global indices $(3,4,5,11,13,16,17,18)$. Candidate pooling,
rollout ranking, verification, and coefficient optimization then used
evidence from all 24 model-construction windows, rather than only the eight
inputs from which candidates were generated.

The long-horizon verifier rollouts began from model-construction initial
conditions 0, 4, and 8 in the left-most table. This reuse was intentional:
the verifiers assess dynamical admissibility from multiple locations in the
model-construction domain and do not provide a held-out performance
estimate. The eight initial conditions in the center table were used only
for the beam--temperature comparison in
Section~\ref{sec:supp-beam-temperature}. The eight initial conditions in the
right-most table were reserved for final testing after the
candidate-generation settings and equations had been fixed. These two
evaluation sets are disjoint from the model-construction bank and from one
another.

\subsection{Selected equations and verifier outcomes}

The canonical system has
$\dot{x}_0=x_1$ and
$\dot{x}_1=\mu x_1-x_0-\mu x_0^2x_1$.
The selected VG equations are written below in the factorized form returned
by the symbolic workflow. Coefficients are rounded only for presentation;
verification and rollout evaluation used the full-precision values.

\noindent\textbf{$\boldsymbol{\mu=0.5}$.}
\begin{equation}
\begin{aligned}
\text{Pre-optimization:}\qquad
  \dot{x}_0
  &=
  1.0239x_1-0.0548x_0,\\
  \dot{x}_1
  &=
  0.2763x_1-1.0463x_0
  -0.0769x_1\!\left(0.0736x_0+3.6009x_0^2\right).
\end{aligned}
\label{eq:supp-vdp-mu05-preopt}
\end{equation}
\begin{equation}
\begin{aligned}
\text{Post-optimization:}\qquad
  \dot{x}_0
  &=
  0.99726x_1-0.012400x_0,\\
  \dot{x}_1
  &=
  0.54166x_1-0.99708x_0
  -0.13251x_1\!\left(0.029451x_0+4.0969x_0^2\right).
\end{aligned}
\label{eq:supp-vdp-mu05-postopt}
\end{equation}

\noindent\textbf{$\boldsymbol{\mu=1.5}$.}
\begin{equation}
\begin{aligned}
\text{Pre-optimization:}\qquad
  \dot{x}_0
  &=
  0.0007\left(1+1.2715x_0\right)^2+0.9889x_1,\\
  \dot{x}_1
  &=
  1.4616x_1-0.9972x_0
  -0.1104x_1\!\left(0.0690x_0+12.5443x_0^2\right).
\end{aligned}
\label{eq:supp-vdp-mu15-preopt}
\end{equation}
\begin{equation}
\begin{aligned}
\text{Post-optimization:}\qquad
  \dot{x}_0
  &=
  0.00067573\left(0.99818+1.2726x_0\right)^2
  +0.99637x_1,\\
  \dot{x}_1
  &=
  1.43998x_1-1.00151x_0
  -0.11351x_1\!\left(0.070367x_0+12.8741x_0^2\right).
\end{aligned}
\label{eq:supp-vdp-mu15-postopt}
\end{equation}

Comparison of each pre- and post-optimization pair shows that coefficient
optimization preserves the generated symbolic structure and changes only
its numerical constants; it does not add or remove terms. For each value of
$\mu$, only one of the ten rollout-ranked candidates passed the complete
pre-optimization verifier suite. That candidate also had the lowest
aggregate rollout error. The verifier suite therefore provided an
independent admissibility requirement without changing the selected
candidate in these two cases.

The selected equations were verified again after coefficient optimization.
Table~\ref{tab:supp-vdp-verifier-results} places each measured outcome beside
the corresponding Van der Pol acceptance requirement. The local
vector-field limit is shortlist-relative and therefore differs between the
two parameter values; the remaining limits are the fixed settings defined in
Section~\ref{sec:supp-verifiers}.

\begin{table}[!ht]
\centering
\small
\caption{Post-optimization verifier outcomes for the selected Van der Pol
equations. Percentages are reported relative to the corresponding
ground-truth oscillation statistic.}
\label{tab:supp-vdp-verifier-results}
\setlength{\tabcolsep}{5pt}
\begin{tabularx}{\linewidth}{
  >{\raggedright\arraybackslash}p{1.30in}
  >{\raggedright\arraybackslash}X
  >{\centering\arraybackslash}p{1.22in}
  >{\centering\arraybackslash}p{1.22in}}
  \toprule
  Verifier & Acceptance requirement
  & $\mu=0.5$ & $\mu=1.5$ \\
  \midrule
  Finite integration
  & All three long-horizon rollouts complete
  & 3/3 & 3/3 \\
  Boundedness
  & Maximum state norm no greater than $1.5$ times the observed maximum
  & Pass & Pass \\
  Local vector field
  & Score no greater than the case-specific limit
  & $8.72{\times}10^{-4}$ \newline (limit $4.02{\times}10^{-2}$)
  & $2.11{\times}10^{-4}$ \newline (limit $4.60{\times}10^{-3}$) \\
  Amplitude error
  & No greater than 15\%
  & 0.741\% & 0.678\% \\
  Period error
  & No greater than 10\%
  & 0.263\% & 0.236\% \\
  Amplitude CV
  & No greater than 10\%
  & 0.057\% & 0.016\% \\
  Period CV
  & No greater than 5\%
  & 0.370\% & Below numerical resolution \\
  \midrule
  Overall outcome
  & Every applicable requirement satisfied
  & Pass & Pass \\
  \bottomrule
\end{tabularx}
\end{table}

The local vector-field scores are 2.17\% and 4.59\% of their respective
limits for $\mu=0.5$ and $1.5$. Across both systems, the amplitude errors
remain below 0.75\%, the period errors below 0.27\%, and the cross-start
coefficients of variation below 0.38\%. The post-optimization equations thus
satisfy the numerical-integration, boundedness, local-dynamics, oscillation,
and attractor-consistency requirements from all three verifier initial
conditions. This reverification establishes dynamical admissibility after
the constants have changed;
Section~\ref{sec:supp-vdp-rollouts} separately assesses predictive
generalization from initial conditions excluded from model construction.

\subsection{Representative held-out rollouts}
\label{sec:supp-vdp-rollouts}

Across the eight final-test initial conditions, coefficient optimization
reduced the median VG rollout NMSE from 0.287 to
$6.99\times10^{-4}$ for $\mu=0.5$, and from 0.0139 to
$3.27\times10^{-3}$ for $\mu=1.5$. These values differ from the validation
errors in Table~\ref{tab:supp-beam-temperature}: the validation errors
selected the candidate-generation settings, whereas the present values were
obtained from the previously unused final-test trajectories after those
settings and the equations had been fixed.

\begin{figure}[!ht]
  \centering
  \includegraphics[width=\linewidth]{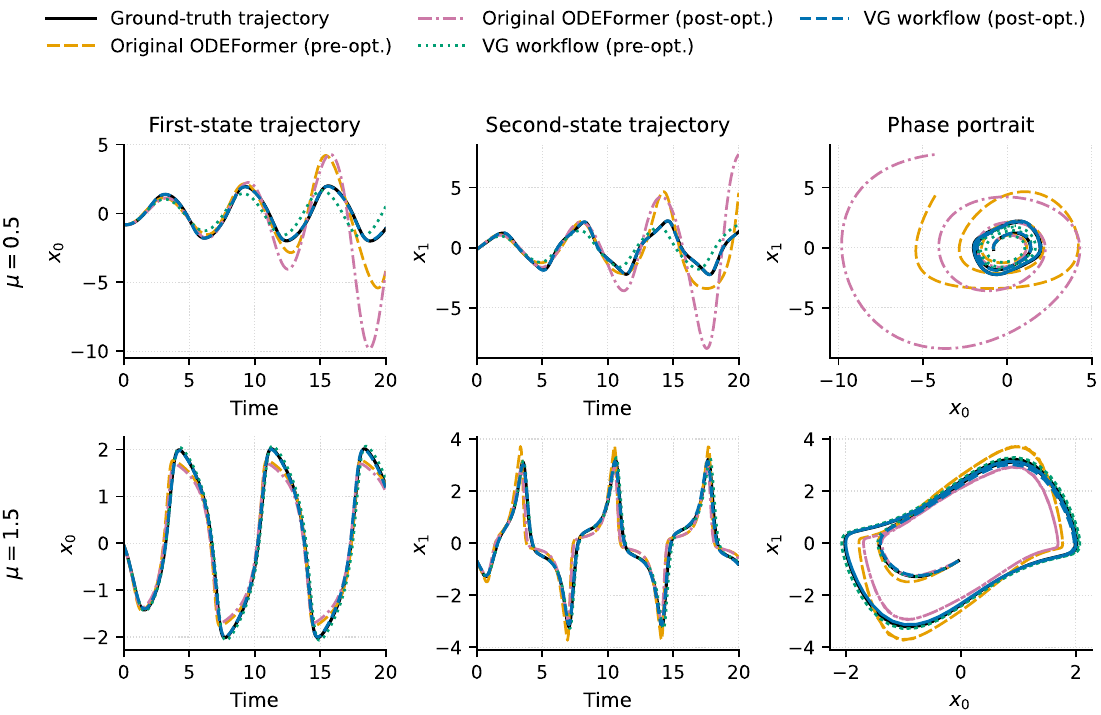}
  \caption{Representative final-test Van der Pol rollouts. Ground-truth
  trajectories are compared with equations from the original
  single-trajectory ODEFormer workflow and the VG workflow, in each case
  before and after coefficient optimization.}
  \label{fig:supp-vdp-rollouts}
\end{figure}

Figure~\ref{fig:supp-vdp-rollouts} shows one representative final-test case
for each parameter. The representative-case rule was defined over the eight
final-test trajectories before inspecting the plots. An eligible trajectory
required complete rollouts from the original ODEFormer equation before
coefficient optimization and from the VG equation both before and after
optimization. Among these eligible trajectories, we selected the case whose
post-optimization VG NMSE was closest to the median over all eight
final-test trajectories. When the two observations bracketing the median
were equally close, the lower trajectory index was used. This rule gives
test trajectory 2, with
$\mathbf{x}(0)=(-0.811743,-0.133746)$, for $\mu=0.5$, and test trajectory 3,
with $\mathbf{x}(0)=(-0.041897,-0.680522)$, for $\mu=1.5$. The
post-optimization original ODEFormer rollout is included in the figure to
complete the before--after comparison but did not influence representative
case selection.

\FloatBarrier

For $\mu=0.5$, both original ODEFormer equations begin to separate markedly
from the ground truth after approximately $t=10$. Their oscillation
amplitudes then grow rapidly, and neither preserves the bounded
ground-truth cycle over the full displayed interval. The pre-optimization VG
equation remains bounded and oscillatory but accumulates a visible phase
discrepancy. Coefficient optimization brings the VG trajectory and phase
portrait into close agreement with the ground truth. For $\mu=1.5$, all four
learned rollouts remain oscillatory. The original ODEFormer equations before
and after optimization trace visibly different cycles, whereas the
post-optimization VG rollout follows the ground-truth cycle throughout the
interval; its smaller residual phase and shape differences are consistent
with the final-test NMSE.

The verifiers therefore contribute more than a check on numerical
solvability: before selection, they require finite and bounded integration,
local vector-field agreement, accurate oscillation amplitude and period, and
consistent attractor statistics across several initial conditions. The
final-test rollouts then provide separate evidence that the selected,
reverified equations preserve the relevant oscillatory dynamics from initial
conditions excluded from model construction. Together with the aggregate
comparison in Figure~\ref{fig:vdp-errors}, these results show that the VG advantage is
supported by explicit multi-start dynamical requirements as well as by
held-out predictive accuracy, rather than by low average rollout error alone.

\section{Fixed-Reynolds-Number Representation and Selection Details}
\label{sec:supp-fixed-re}

\noindent
This section expands the fixed-$\Rey=300$ experiment reported in Section~\ref{sec:fixed-re-results}.
After the initial 500 snapshots were discarded, the remaining time series was
divided chronologically into a 250-snapshot development interval, a
100-snapshot validation interval, and a 150-snapshot final-test interval. The
development interval supplied the POD representation and supported candidate
generation, multi-window ranking and verification, and coefficient
optimization. The validation interval was used to choose the POD rank and
decoding-window configuration; the final-test interval remained held out
until those choices were complete. Sections~\ref{sec:supp-fixed-pod},
\ref{sec:supp-fixed-re-matrix}, and \ref{sec:supp-fixed-test} respectively
examine the representation, configuration selection, and final-test
behavior.

\subsection{POD spectrum and mode-pair diagnostics}
\label{sec:supp-fixed-pod}

POD was applied to fluctuations about the development-interval mean, rather
than to the total vorticity fields. If
$\boldsymbol{\omega}(t_i)$ is the vectorized field at development time $t_i$ and
$\overline{\boldsymbol{\omega}}_{\!D}$ is their temporal mean, the fluctuation
snapshot matrix and its singular-value decomposition are
\begin{equation}
\begin{aligned}
  \mathbf X'
  &=
  \begin{bmatrix}
    \boldsymbol{\omega}(t_1)-\overline{\boldsymbol{\omega}}_{\!D}
    & \cdots &
    \boldsymbol{\omega}(t_{N_D})-\overline{\boldsymbol{\omega}}_{\!D}
  \end{bmatrix}
  =
  \mathbf U\boldsymbol{\Sigma}\mathbf V^{\mathsf T},\\
  E_r
  &=
  \frac{\sum_{j=1}^{r}\sigma_j^2}
       {\sum_{j=1}^{N_s}\sigma_j^2}.
\end{aligned}
\label{eq:supp-pod-energy}
\end{equation}
Here $N_D=250$, $\sigma_j$ is the $j$th singular value, $N_s$ is the number
of nonzero singular values, and $E_r$ is the fraction of fluctuation energy
represented by the first $r$ modes. The mean field is restored only when the
reduced coordinates are reconstructed in physical space.

\begin{figure}[!ht]
  \centering
  \makebox[\linewidth][c]{%
    \includegraphics[width=1.02\linewidth]{
      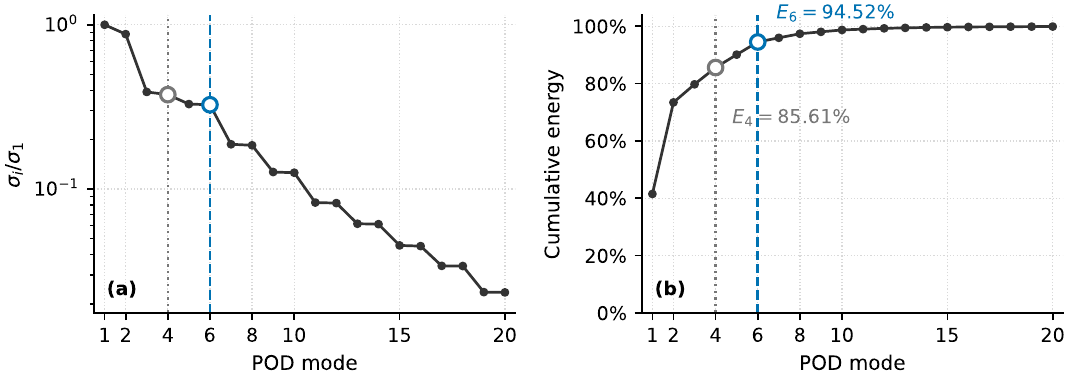}}
  \caption{POD singular-value spectrum and cumulative fluctuation energy at
  fixed $\Rey=300$.}
  \label{fig:supp-fixed-pod-spectrum}
\end{figure}

The near-equal singular values in Figure~\ref{fig:supp-fixed-pod-spectrum}
form three leading pairs, as expected for quadrature components of an
oscillatory wake. We therefore compared two even ranks that preserve complete
pairs: rank four gives a compact representation of the first two pairs,
whereas rank six adds the third pair and retains a larger fraction of the
wake fluctuations. This comparison tests whether the additional oscillatory
content improves field reconstruction sufficiently to justify a
higher-dimensional symbolic system. Ranks four and six retain 85.61\% and
94.52\% of the fluctuation energy, respectively.

For a consecutive pair $(2k-1,2k)$, we quantify its relative energy imbalance
by
\begin{equation}
  \delta_k
  =
  \frac{|\sigma_{2k-1}^{2}-\sigma_{2k}^{2}|}
       {\sigma_{2k-1}^{2}+\sigma_{2k}^{2}}.
  \label{eq:supp-pod-pair-imbalance}
\end{equation}
The pair-energy column in Table~\ref{tab:supp-pod-pairs} is the percentage of
the total fluctuation energy carried by the two modes in that row; the
cumulative column gives the energy retained through that pair. The first pair
therefore accounts for 73.44\% of the wake fluctuations. The second adds
12.17 percentage points, bringing the rank-four representation to 85.61\%,
and the third adds a further 8.91 percentage points, bringing rank six to
94.52\%. Their coefficient trajectories vary on successively faster time
scales, corresponding to the fundamental shedding oscillation and its higher
harmonics. The imbalance $\delta_k$ equals zero for an exactly energy-matched
pair and increases as one member becomes dominant. Its small and decreasing
values show that each retained pair uses both quadrature components, most
closely for modes 5--6. Table~\ref{tab:supp-pod-pairs} therefore complements
the cumulative spectrum by separating the energy contributed by each
oscillatory pair.

\begin{table}[!ht]
  \centering
  \caption{POD mode-pair energy and imbalance diagnostics.}
  \label{tab:supp-pod-pairs}
  \begin{tabular}{lrrr}
    \toprule
    Modes & Pair energy (\%) & Cumulative energy (\%) & $\delta_k$ \\
    \midrule
    1--2 & 73.44 & 73.44 & 0.1309 \\
    3--4 & 12.17 & 85.61 & 0.0379 \\
    5--6 & 8.91  & 94.52 & 0.0097 \\
    \bottomrule
  \end{tabular}
\end{table}
\FloatBarrier

\subsection{POD-rank and decoding-window selection matrix}
\label{sec:supp-fixed-re-matrix}

The representation and decoding choices motivated in
Section~\ref{sec:supp-fixed-pod} were compared without using the final-test
interval. For each rank, the VG workflow generated candidates independently
from either eight evenly distributed windows or all 24 windows in the
development interval. The eight-window setting tests whether broad temporal
coverage is sufficient without decoding every available window; the
24-window setting instead maximizes the diversity of generated candidates.
The resulting pools contained 160 and 480 equation systems, respectively.
In every configuration, candidate ranking, verification, and coefficient
optimization used all 24 windows. Only the number of windows supplied
independently to the symbolic backbone during candidate generation was
varied.

Table~\ref{tab:supp-fixed-re-matrix} reports the validation comparison. The
rank-four models have the smallest coordinate-space errors, but their POD
truncation error exceeds 20\%, and their end-to-end field errors consequently
remain above 21\%. Rank six reduces the representation error to 12.70\%.
Within this rank, decoding from eight windows gives the smaller continuous
rollout and end-to-end field errors. It was therefore chosen before the
final-test interval was evaluated.

\begin{table}[!ht]
  \centering
  \setlength{\belowcaptionskip}{5pt}
  \caption{Fixed-$\Rey$ validation matrix. ``Window'' and ``Continuous'' are
  median coordinate errors; POD, VG--POD, and end-to-end are mean relative
  field errors. All errors are percentages, and ``Adm.'' gives the number of
  verifier-admissible candidates in the ten-candidate shortlist.}
  \label{tab:supp-fixed-re-matrix}
  \small
  \begin{tabular}{rrrccccc}
    \toprule
    Rank & Windows & Adm. & Window & Continuous & POD & VG--POD & End-to-end \\
    \midrule
    4 & 8  & 10 & 1.30 & 1.54 & 20.87 & 4.57  & 21.39 \\
    4 & 24 & 10 & 1.11 & 2.03 & 20.87 & 5.88  & 21.75 \\
    6 & 8  & 6  & 4.36 & 3.95 & 12.70 & 7.57  & 14.93 \\
    6 & 24 & 10 & 4.61 & 7.11 & 12.70 & 12.84 & 18.28 \\
    \bottomrule
  \end{tabular}
\end{table}

Table~\ref{tab:supp-fixed-re-matrix} reports the admissible counts obtained at
the nominal verifier settings. Section~\ref{sec:supp-tolerance-sensitivity}
examines how these counts change when the quantitative agreement tolerances
are tightened or relaxed. For the selected rank-six, eight-window
configuration, Eq.~\eqref{eq:supp-fixed-selected-preopt} below is the
pre-optimization equation selected at the nominal settings, and
Eq.~\eqref{eq:supp-fixed-alternative} is the next-lowest-rollout admissible
alternative. These are the two equations referenced in
Section~\ref{sec:supp-tolerance-sensitivity}. Although the number of
admissible rank-six candidates changes across the tested tolerance factors,
Eq.~\eqref{eq:supp-fixed-selected-preopt} remains selected in every case.

Let $\mathbf z=(z_1,\ldots,z_6)^{\mathsf T}$ denote the standardized POD
coordinates, consistent with the notation used for the post-optimization
system in Eq.~\eqref{eq:fixed-re-system}. The pre-optimization system with the lowest development-
window rollout error, whose structure was advanced to coefficient
optimization, was
\begin{equation}
\begin{aligned}
  \dot z_1 &= 1.0044z_2-0.0700(12.3400-0.9594z_2)^{-1},&
  \dot z_2 &= -1.1105z_1,\\
  \dot z_3 &= -2.1602z_4,&
  \dot z_4 &= 2.1867z_3,\\
  \dot z_5 &= 3.4432z_6-0.1382\sin(0.1052+9.1998z_2),&
  \dot z_6 &= -0.1399-2.8694z_5.
\end{aligned}
\label{eq:supp-fixed-selected-preopt}
\end{equation}
The next-lowest-rollout admissible system provides a representative
alternative:
\begin{equation}
\begin{aligned}
  \dot z_1 &= 1.1208z_2,&
  \dot z_2 &= -1.0604z_1,\\
  \dot z_3 &= -2.2995z_4,&
  \dot z_4 &= 2.0672z_3,\\
  \dot z_5 &= 3.1395z_6-0.2202\sin(11.8700+148.5203z_6),&
  \dot z_6 &= -0.1463-3.1228z_5.
\end{aligned}
\label{eq:supp-fixed-alternative}
\end{equation}
Both systems preserve three oscillator pairs and satisfy all applicable
verifiers, but their nonlinear structures differ. In
Eq.~\eqref{eq:supp-fixed-selected-preopt},
$-0.0700(12.3400-0.9594z_2)^{-1}$ adds a reciprocal correction to
$\dot z_1$ in the first pair, while
$-0.1382\sin(0.1052+9.1998z_2)$ couples the first pair to $\dot z_5$ in
the third pair. Equation~\eqref{eq:supp-fixed-alternative} leaves the first
pair linear and instead introduces
$-0.2202\sin(11.8700+148.5203z_6)$ as a nonlinear correction within the
third pair. Both systems also contain a small constant offset in $\dot z_6$.

At the nominal settings, six of the ten shortlisted systems for this
configuration satisfy every verifier. The coexistence of
Eqs.~\eqref{eq:supp-fixed-selected-preopt} and
\eqref{eq:supp-fixed-alternative} therefore illustrates the practical
non-uniqueness discussed in Section~\ref{sec:supp-identifiability}:
verification excludes four rollout-competitive candidates but does not imply
a unique symbolic structure within the admissible set. Rollout error provides
the remaining selection criterion and chooses
Eq.~\eqref{eq:supp-fixed-selected-preopt}. This supports improved
discrimination within the generated hypothesis class, rather than global
structural identifiability. Coefficient optimization of the selected
equation produced the post-optimization system reported in Eq.~\eqref{eq:fixed-re-system}.
\FloatBarrier

\subsection{Extended coordinate and field comparisons}
\label{sec:supp-fixed-test}

\begin{figure}[!ht]
  \centering
  \includegraphics[width=\linewidth]{
    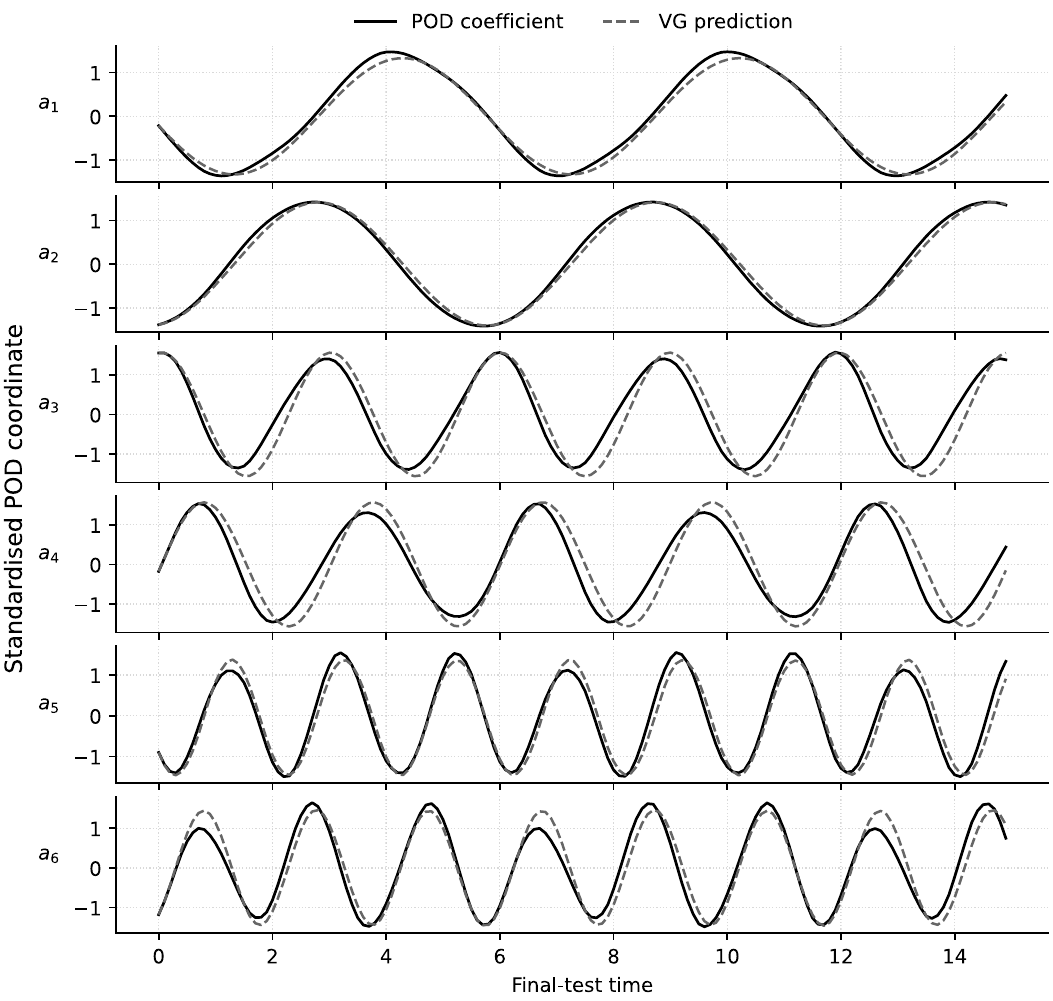}
  \caption{Final-test fixed-$\Rey$ rollouts in the six standardized POD
  coordinates.}
  \label{fig:supp-fixed-test-rollouts}
\end{figure}

\begin{figure}[!ht]
  \centering
  \includegraphics[width=\linewidth]{
    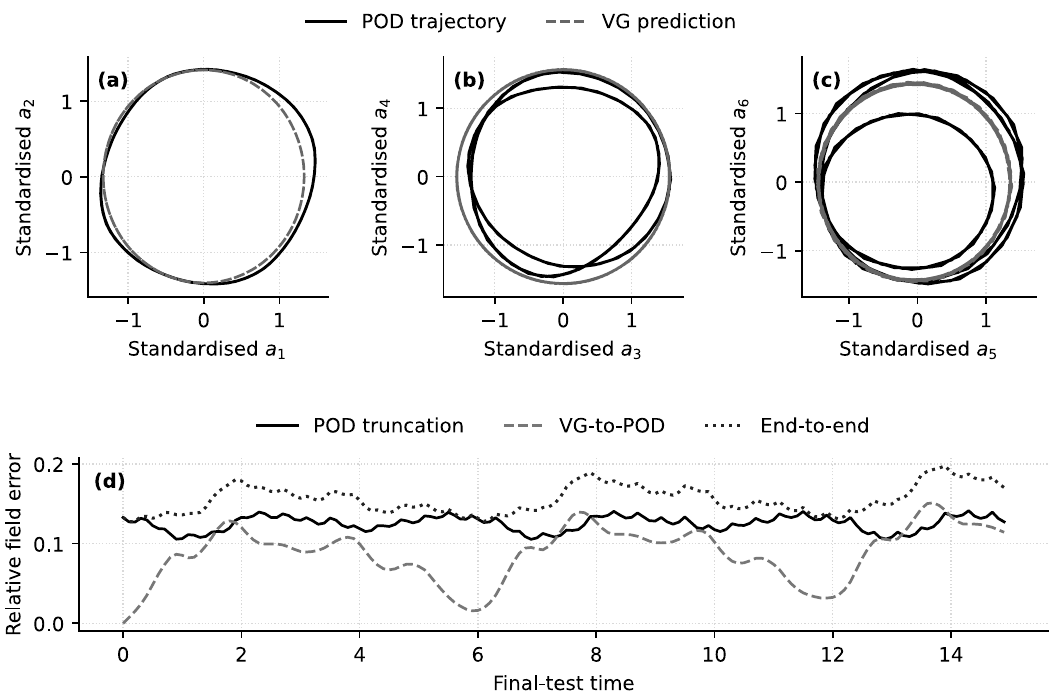}
  \caption{Final-test fixed-$\Rey$ comparisons: phase portraits for
  (a) $a_1$--$a_2$, (b) $a_3$--$a_4$, and (c) $a_5$--$a_6$; (d) POD
  truncation, VG-to-POD, and end-to-end field errors.}
  \label{fig:supp-fixed-test-phase-errors}
\end{figure}

The final-test interval was evaluated only after the rank-six, eight-window
configuration had been chosen. In Figures~\ref{fig:supp-fixed-test-rollouts}
and \ref{fig:supp-fixed-test-phase-errors}, $a_j$ denotes the plotted
standardized POD coefficient and corresponds to $z_j$ in
Eqs.~\eqref{eq:supp-fixed-selected-preopt} and
\eqref{eq:supp-fixed-alternative}. The first oscillator pair remains closely
aligned throughout the interval. The second pair develops the most visible
phase difference, while the third retains the higher-frequency oscillation
with moderate amplitude and phase differences. The phase portraits in
Figure~\ref{fig:supp-fixed-test-phase-errors}(a)--(c) show that all three
predicted pairs remain bounded and continue to trace closed oscillatory
paths. The first predicted orbit remains close to its POD counterpart; the
higher-harmonic pairs show greater differences in orbit geometry. Because
these pairs carry substantially less fluctuation energy than the first pair,
their coordinate-level differences have a smaller influence on the
reconstructed vorticity field.

Field error was decomposed into three contributions. For the simulated
vorticity field $\boldsymbol{\omega}(t)$, its rank-six POD projection
$\boldsymbol{\omega}_{\mathrm{POD}}(t)$, and the field reconstructed from the
symbolic rollout $\widehat{\boldsymbol{\omega}}_{\mathrm{VG}}(t)$, the
instantaneous relative errors are
\begin{equation}
\begin{aligned}
 e_{\mathrm{POD}}(t)
 &=
 \frac{\|\boldsymbol{\omega}_{\mathrm{POD}}(t)
       -\boldsymbol{\omega}(t)\|_2}
      {\|\boldsymbol{\omega}(t)\|_2},\\
 e_{\mathrm{VG\text{-}POD}}(t)
 &=
 \frac{\|\widehat{\boldsymbol{\omega}}_{\mathrm{VG}}(t)
       -\boldsymbol{\omega}_{\mathrm{POD}}(t)\|_2}
      {\|\boldsymbol{\omega}_{\mathrm{POD}}(t)\|_2},\\
 e_{\mathrm{end}}(t)
 &=
 \frac{\|\widehat{\boldsymbol{\omega}}_{\mathrm{VG}}(t)
       -\boldsymbol{\omega}(t)\|_2}
      {\|\boldsymbol{\omega}(t)\|_2}.
\end{aligned}
\label{eq:supp-fixed-field-errors}
\end{equation}
Their time-averaged values are 12.53\%, 8.77\%, and 15.56\%,
respectively. Figure~\ref{fig:supp-fixed-test-phase-errors}(d) shows their
variation over the final-test interval. All three errors remain bounded
without systematic growth. The smaller VG-to-POD error indicates that the
symbolic rollout remains closer to the retained six-mode trajectory than the
POD reconstruction is to the simulation. The end-to-end error reflects both
the POD truncation and the symbolic-rollout discrepancy, but it is not their
arithmetic sum because the error fields are neither collinear nor normalized
by the same reference.

\begin{figure}[!ht]
  \centering
  \includegraphics[width=\linewidth]{
    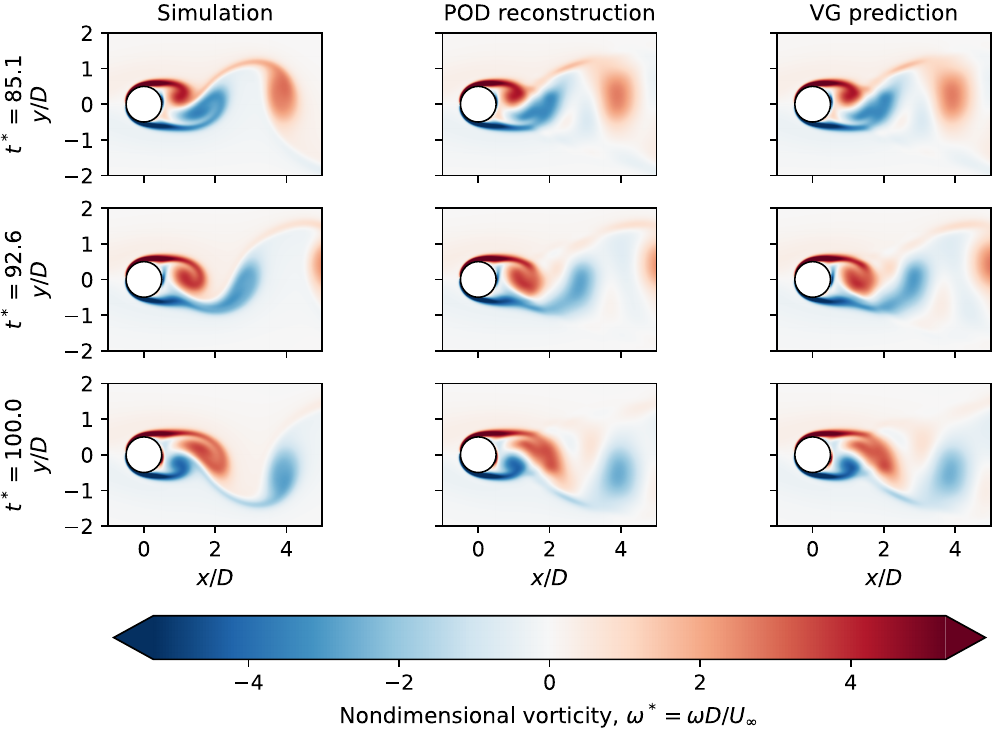}
  \caption{Simulation, rank-six POD reconstruction, and VG prediction at
  three times in the final-test fixed-$\Rey$ interval.}
  \label{fig:supp-fixed-test-fields}
\end{figure}

Representative vorticity fields are shown in
Figure~\ref{fig:supp-fixed-test-fields}. The POD reconstruction preserves the
alternating vortex street but smooths the smaller-scale vorticity gradients
omitted by the six-mode representation. At all three times, the VG
reconstruction closely follows the shedding phase and large-scale
organization of the POD field. Its remaining deviation from the simulation
reflects the POD truncation visible in the middle column together with the
smaller additional differences introduced by the symbolic coordinate
rollout.

\FloatBarrier

\section{Autoencoder Representation--Discoverability Sensitivity}
\label{sec:supp-ae-sensitivity}

\subsection{Controlled architectures, seeds, and optimization protocol}

Unlike POD coordinates, which follow directly from a covariance decomposition
of a specified snapshot set up to sign and rotations within degenerate
subspaces, nonlinear autoencoders do not define a unique latent embedding.
Different architectures, initializations, and optimization paths may yield
similar field reconstructions while producing latent trajectories with
different temporal geometry. This non-uniqueness motivates examining
representation fidelity together with properties relevant to symbolic
discoverability.

\begin{figure}[!ht]
  \centering
  \includegraphics[width=0.96\linewidth]{
    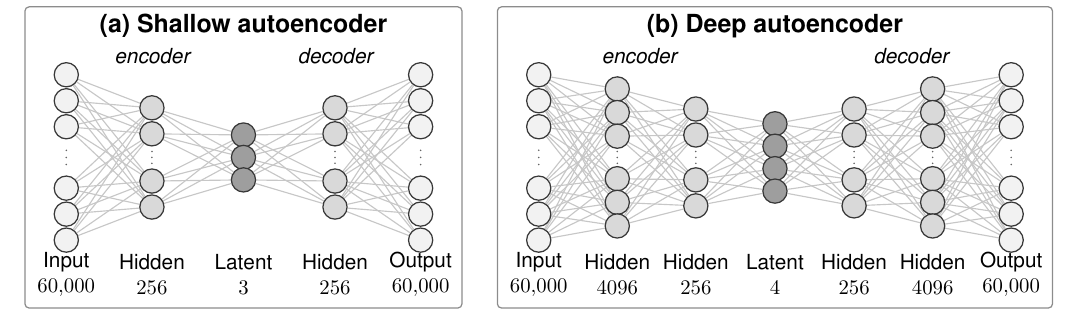}
  \caption{Fully connected shallow and deep autoencoder architectures used in
  the representation study. Hidden layers use ReLU activation; latent and
  output layers are linear.}
  \label{fig:supp-ae-architectures}
\end{figure}

The controlled representation study crossed latent dimensions three and four
with shallow and deep fully connected autoencoders. The shallow encoder used
one hidden layer of width 256, whereas the deep encoder used hidden widths
4096 and 256; both used symmetric decoders, as shown in
Figure~\ref{fig:supp-ae-architectures}. The independently prepared
autoencoder used for the cross-$\Rey$ result in Section~\ref{sec:multi-re-results} has the same
architecture as the shallow three-coordinate autoencoders tested here. It
was trained before the controlled study without a prespecified, recorded
initialization seed and is therefore retained as a separate comparison
rather than treated as an additional controlled seed.

\FloatBarrier

Each configuration was initialized from seeds 17, 29, and 43 and optimized
using the seven Reynolds numbers employed for symbolic model construction and
the same 500 post-transient snapshots per Reynolds number. All comparisons in
Table~\ref{tab:supp-ae-representation-sensitivity} use these post-transient
fields. One predeclared seed per architecture was carried forward to the
symbolic discovery comparison.

\subsection{Definition of the latent-representation diagnostics}

Three diagnostics characterize the temporal properties of each embedding before
symbolic model construction. Let
$\mathbf Z^{(m)}\in\mathbb R^{T\times d_z}$ contain the latent trajectory at
the $m$th Reynolds number, where $T$ is the number of snapshots and
$d_z$ the latent dimension. Each coordinate is standardized using its mean
$\mu_j$ and standard deviation $\sigma_j$ on the concatenated trajectories:
$\widetilde Z_{t,j}^{(m)}=(Z_{t,j}^{(m)}-\mu_j)/(\sigma_j+\epsilon)$.
With $\Delta$ and $\Delta^2$ denoting first and second temporal differences,
respectively, the latent roughness is
\begin{equation}
  \mathcal R_z
  =
  \frac{1}{M_R}\sum_{m=1}^{M_R}
  \frac{
    \operatorname{mean}_{t,j}
    \left[(\Delta^2\widetilde Z_{t,j}^{(m)})^2\right]
  }{
    \operatorname{mean}_{t,j}
    \left[(\Delta\widetilde Z_{t,j}^{(m)})^2\right]+\epsilon
  },
  \label{eq:latent-roughness}
\end{equation}
where $M_R=7$ is the number of Reynolds numbers. The denominator
normalizes by the overall first-difference scale, so
Eq.~\eqref{eq:latent-roughness} measures rapid changes in temporal slope
rather than latent amplitude. Smaller values indicate smoother trajectories.

For spectral concentration, let $P_{k,j}^{(m)}$ be the discrete Fourier power
of the centered standardized coordinate at positive, nonzero frequency bin
$k$, and let $\mathcal K_3^{(m,j)}$ contain its three largest-power bins. The
reported diagnostic is
\begin{equation}
  \mathcal S_z
  =
  \frac{1}{M_Rd_z}
  \sum_{m=1}^{M_R}\sum_{j=1}^{d_z}
  \frac{
    \sum_{k\in\mathcal K_3^{(m,j)}}P_{k,j}^{(m)}
  }{
    \sum_{k>0}P_{k,j}^{(m)}+\epsilon
  }.
  \label{eq:latent-spectral-concentration}
\end{equation}
Values of $\mathcal S_z$ in
Eq.~\eqref{eq:latent-spectral-concentration} closer to one indicate that most
temporal variation is organized around a small number of frequencies, as
expected for a dominant shedding oscillator and its harmonics.

Finally, concatenate the unstandardized trajectories into
$\mathbf Z_{\mathrm{all}}$ and let
$\lambda_1,\ldots,\lambda_{d_z}\geq0$ be the eigenvalues of its sample
covariance matrix. The covariance participation ratio is
\begin{equation}
  d_{\mathrm{PR}}
  =
  \frac{
    \left(\sum_{j=1}^{d_z}\lambda_j\right)^2
  }{
    \sum_{j=1}^{d_z}\lambda_j^2+\epsilon
  }.
  \label{eq:latent-participation}
\end{equation}
This effective dimension equals one when nearly all covariance is concentrated
in one direction and increases as variance is distributed more evenly among
latent coordinates. Equations~\eqref{eq:latent-roughness}--\eqref{eq:latent-participation}
describe complementary properties: temporal regularity, spectral
organization, and covariance use of the available latent dimensions.

\subsection{Latent-representation diagnostics evaluation}

Table~\ref{tab:supp-ae-representation-sensitivity} compares reconstruction
fidelity with the three latent diagnostics. The smallest reconstruction
error, 4.494\%, was attained by the deep three-coordinate model initialized
with seed 43. This representation did not have the smoothest latent
trajectories: its roughness was 0.0465, compared with 0.0224 for the
independent shallow reference. The latter also had the largest spectral
concentration, 0.9264, despite its higher reconstruction error of 5.826\%.
The comparison therefore separates field reconstruction from temporal
properties that affect compatibility with the symbolic backbone. As the
symbolic-discovery outcomes in
Table~\ref{tab:supp-ae-symbolic-outcomes} subsequently show, an accurate
snapshot reconstruction need not yield latent dynamics that are readily
expressed within the backbone's learned symbolic distribution.

\begin{table}[!ht]
  \centering
  \caption{Autoencoder representation sensitivity across architecture and
  initialization seed. Field error is the temporal mean of snapshot-wise
  relative spatial $L_2$ error; the latent diagnostics are defined in
  Eqs.~\eqref{eq:latent-roughness}--\eqref{eq:latent-participation}.}
  \label{tab:supp-ae-representation-sensitivity}
  \small
  \begin{tabular}{lrrrrr}
    \toprule
    Configuration & Seed & Field error (\%) & Roughness &
    Spectral conc. & Participation \\
    \midrule
    AE3S & 17 & 5.636 & 0.0472 & 0.8830 & 1.975 \\
    AE3S & 29 & 6.349 & 0.0543 & 0.8841 & 1.910 \\
    AE3S & $43^{*}$ & 5.637 & 0.0332 & 0.8970 & 2.447 \\
    AE4S & $17^{*}$ & 5.379 & 0.1145 & 0.7842 & 1.733 \\
    AE4S & 29 & 5.825 & 0.1082 & 0.8065 & 2.664 \\
    AE4S & 43 & 5.021 & 0.1143 & 0.7877 & 2.854 \\
    AE3D & $17^{*}$ & 4.838 & 0.0347 & 0.9232 & 2.291 \\
    AE3D & 29 & 5.644 & 0.0383 & 0.9087 & 2.370 \\
    AE3D & 43 & 4.494 & 0.0465 & 0.9098 & 2.562 \\
    AE4D & 17 & 5.091 & 0.0993 & 0.8646 & 3.677 \\
    AE4D & $29^{*}$ & 4.625 & 0.0935 & 0.8753 & 3.204 \\
    AE4D & 43 & 4.650 & 0.0733 & 0.8759 & 3.650 \\
    \midrule
    Independent shallow reference$^{**}$ & -- & 5.826 & 0.0224 & 0.9264 & 2.149 \\
    \bottomrule
  \end{tabular}
  \parbox{\linewidth}{\footnotesize\raggedright
  $^{*}$Seed carried forward for the corresponding controlled architecture.
  \par\smallskip
  $^{**}$The independent shallow reference is the separately prepared
  autoencoder used for the cross-parameter result reported in Section~\ref{sec:multi-re-results}. It
  has the same architecture as the shallow three-coordinate autoencoders but
  no prespecified, recorded initialization seed and is not one of the 12
  controlled runs.}
\end{table}
\FloatBarrier

\subsection{Corresponding symbolic-discovery outcomes}

The same beam size of 20 and sampling temperature of 0.1 were used for the
controlled symbolic-discovery comparison and for the independent shallow
reference. Among the four controlled encodings identified by asterisks in
Table~\ref{tab:supp-ae-representation-sensitivity}, only AE3D produced a
pre-optimization candidate satisfying every verifier. Its multi-trajectory
rollout loss was 2.169; the post-optimization system was not admissible. By
contrast,
the independently prepared shallow reference used for the cross-parameter
result in Section~\ref{sec:multi-re-results} yielded an admissible pre-optimization candidate with
loss 1.042, and post-optimization reduced the loss to 0.200 while preserving
admissibility. These outcomes are summarized in
Table~\ref{tab:supp-ae-symbolic-outcomes}.

\begin{table}[!ht]
  \centering
  \caption{Symbolic-discovery outcomes for the representations advanced from
  the controlled study and for the independent shallow reference.}
  \label{tab:supp-ae-symbolic-outcomes}
  \begin{tabular}{lccc}
    \toprule
    Representation & Pre-opt. admissible & Pre-opt. loss & Post-opt. result \\
    \midrule
    AE3S, seed 43 & No & -- & Inadmissible \\
    AE4S, seed 17 & No & -- & Inadmissible \\
    AE3D, seed 17 & Yes & 2.169 & Inadmissible \\
    AE4D, seed 29 & No & -- & Inadmissible \\
    Independent shallow reference & Yes & 1.042 & 0.200, admissible \\
    \bottomrule
  \end{tabular}
\end{table}
\FloatBarrier

The comparison supports a specific conclusion: reconstruction fidelity alone
did not determine whether VG could identify an admissible cross-parameter
equation. It does not imply that roughness or spectral concentration is by
itself sufficient for discoverability; these diagnostics describe relevant
representation properties rather than a complete selection criterion. This
result further supports the joint coordinate-and-equation discovery direction
discussed in Section~\ref{sec:supp-joint-discovery}, in which dynamical
verification can help distinguish latent embeddings with similar
reconstruction fidelity but different compatibility with symbolic discovery.

\section{Extended Results for the Cross-Parameter Cylinder-Flow Case}
\label{sec:supp-multi-re}

\subsection{Representation and evaluation protocol}

The cross-$\Rey$ study used the same shallow autoencoder as the result
reported in Section~\ref{sec:multi-re-results}; Section~\ref{sec:supp-ae-sensitivity} examined this
representation further. The autoencoder operates on mean-subtracted
(fluctuation) vorticity fields. Its three latent coordinates were centered and
scaled using the seven Reynolds numbers employed for symbolic model
construction,
\[
  \Rey\in\{150,200,250,300,350,400,450\}.
\]
The Reynolds number was appended as $r=(\Rey-300)/150$. The symbolic backbone
was given the uniformly sampled trajectories at unit time increments, so its
model time is $\tau=t^*/0.1$, where consecutive physical snapshots are
separated by $\Delta t^*=0.1$. This rescaling changes the numerical values of
the equation coefficients but not the represented trajectories. At each
Reynolds number, the model state was therefore
$\boldsymbol{\xi}=(z_1,z_2,z_3,r)^\mathsf{T}$.

For every Reynolds number, the first 500 simulation snapshots were removed as
the wake-development transient. Model construction and evaluation use the
subsequent 500 post-transient snapshots, spanning approximately
$50\leq t^*\leq100$. The withheld cases comprise interpolation at
$\Rey=175,275,$ and $425$ and extrapolation at $\Rey=500$. None of these four
latent trajectories contributed to generation, ranking, verification, or
coefficient optimization of the symbolic model. The three interpolation
values were also excluded when fitting the autoencoder. The autoencoder was
fitted using fields at $\Rey=500$, but the corresponding latent trajectory
was excluded from symbolic model construction. Thus, ``extrapolation'' at
$\Rey=500$ refers specifically to evaluating the Reynolds-conditioned
equation at $r=4/3$, beyond its construction interval $-1\leq r\leq1$; it
does not refer to extrapolation of the field representation.

Candidates generated from these seven trajectories were verified using the
cross-$\Rey$ settings in Table~\ref{tab:supp-verifier-settings}. The
two-branch shortlist contained 17 distinct systems, of which four satisfied
all applicable verifiers. Coefficient optimization was then applied to the
admissible candidates, followed by the same verifier suite. The resulting
system was chosen before the withheld Reynolds numbers were examined. To
retain the notation used elsewhere in the supplement, dots in
Eqs.~\eqref{eq:supp-multi-re-preopt} and
\eqref{eq:supp-multi-re-postopt} denote differentiation with respect to the
model time $\tau$. The pre-optimization and post-optimization systems are
\begin{align}
 \dot z_1
 &=0.1070z_3-0.0013r+0.0089z_2z_3,
 &
 \dot z_2
 &=0.1003z_1,
 \nonumber\\
 \dot z_3
 &=-0.1098z_1,
 &
 \dot r
 &=0,
 \label{eq:supp-multi-re-preopt}\\[3pt]
 \dot z_1
 &=0.114317z_3+0.004217r+0.020148z_2z_3,
 &
 \dot z_2
 &=0.079842z_1,
 \nonumber\\
 \dot z_3
 &=-0.099544z_1,
 &
 \dot r
 &=0.
 \label{eq:supp-multi-re-postopt}
\end{align}
Since one unit of $\tau$ corresponds to $0.1$ units of $t^*$, multiplying the
right-hand sides of Eq.~\eqref{eq:supp-multi-re-postopt} by ten gives the
physical-time equation reported in Eq.~\eqref{eq:multi-re-system}.

\subsection{Latent-coordinate rollouts at withheld Reynolds numbers}
\label{sec:supp-multi-re-latent}

\begin{figure}[p]
  \centering
  \includegraphics[width=\linewidth]{
    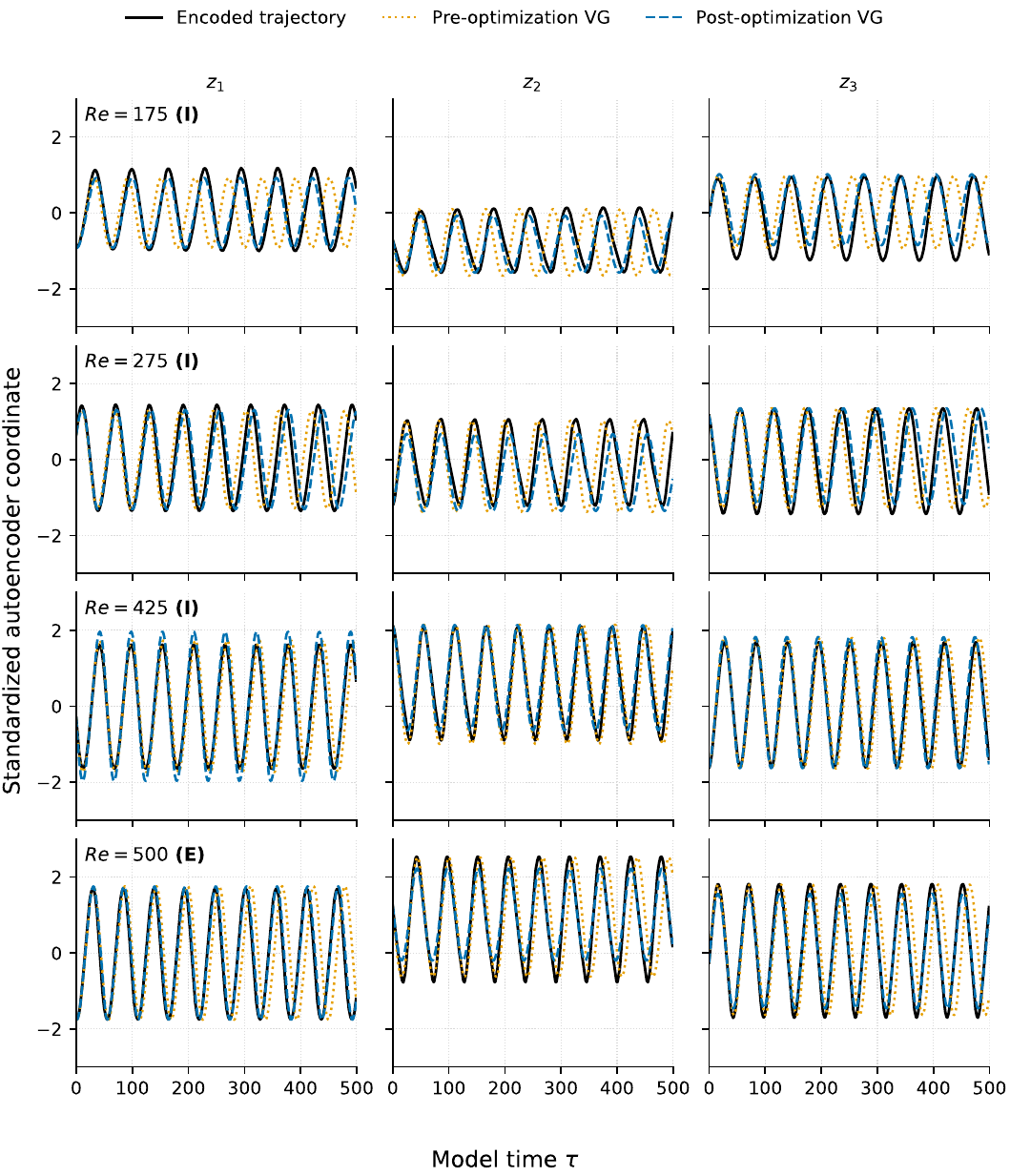}
  \caption{Autoencoder-encoded trajectories and the corresponding
  pre-optimization and post-optimization VG rollouts at the withheld Reynolds
  numbers. I denotes interpolation within the symbolic model-construction
  range, and E denotes extrapolation beyond that range.}
  \label{fig:supp-multi-re-latent}
\end{figure}

Figure~\ref{fig:supp-multi-re-latent} compares the latent trajectories obtained
by encoding the simulated fields with the pre-optimization and
post-optimization VG rollouts. We retain $z_j$ for these nonlinear
autoencoder coordinates, consistent with the notation in
Section~\ref{sec:multi-re-results} and distinct from the POD coefficients $a_j$ in
Figures~\ref{fig:supp-fixed-test-rollouts} and
\ref{fig:supp-fixed-test-phase-errors}. For an encoded trajectory
$\mathbf z(t_n)$ and its prediction $\widehat{\mathbf z}(t_n)$, the latent
normalized mean-squared error is
\begin{equation}
  \mathcal E_z
  =
  \frac{1}{3}\sum_{j=1}^{3}
  \frac{
    N^{-1}\sum_{n=1}^{N}
    \left(\widehat z_j(t_n)-z_j(t_n)\right)^2
  }{
    \operatorname{Var}_{n}\!\left[z_j(t_n)\right]+\epsilon
  },
  \label{eq:supp-multi-re-latent-nmse}
\end{equation}
where $N=500$ and $\epsilon$ is a small numerical constant. This metric
weights each coordinate relative to its own variation; the conserved
parameter coordinate is not included.

Coefficient optimization reduces $\mathcal E_z$ at every withheld Reynolds
number:
\begin{itemize}
  \item $\Rey=175$: $2.312\rightarrow0.142$, a 93.9\% reduction;
  \item $\Rey=275$: $0.797\rightarrow0.247$, a 69.0\% reduction;
  \item $\Rey=425$: $0.188\rightarrow0.043$, a 77.2\% reduction;
  \item $\Rey=500$: $0.714\rightarrow0.036$, a 94.9\% reduction.
\end{itemize}
All post-optimization rollouts remain finite and oscillatory over the
complete interval. The largest remaining error occurs at $\Rey=275$, where a
progressive phase difference is visible in all three coordinates without a
loss of the oscillatory regime. The error ordering is not monotonic with
distance from the Reynolds numbers used for model construction. It also
depends on how accurately the selected equation represents the variation of
latent frequency, phase, and amplitude across the parameter range, and on
the accumulation of small frequency differences over a long rollout. In
Eq.~\eqref{eq:supp-multi-re-postopt}, $r$ enters explicitly through an
additive term in $\dot z_1$, while the oscillator couplings are shared across
Reynolds number. The model can therefore represent a Reynolds-dependent
shift of the common oscillator, but it does not allow every coupling or
frequency coefficient to vary independently with $\Rey$. The strong result
at $\Rey=500$ indicates that the post-optimization equation continues the
same periodic shedding dynamics beyond the parameter range used to construct
it.

\FloatBarrier

\subsection{Spatially averaged wake evolution}

The latent comparison can be related to an observable quantity in the
physical field by averaging the nondimensional vorticity over the wake
region
\[
  \mathcal W
  =
  \left\{(x/D,y/D):0.5\leq x/D\leq5,\;
  |y/D|\leq1.5\right\}.
\]
For a field $\omega^*(\mathbf x,t^*)=\omega D/U_\infty$, its spatial average
is
\begin{equation}
  \left\langle\omega^*\right\rangle_{\mathcal W}(t^*)
  =
  \frac{1}{|\mathcal W|}
  \int_{\mathcal W}\omega^*(\mathbf x,t^*)\,\mathrm d\mathbf x.
  \label{eq:supp-multi-re-spatial-average}
\end{equation}
This integral measures the net balance of positive and negative vorticity
within the specified region and provides a scalar record of shedding phase
and amplitude.

\begin{figure}[!ht]
  \centering
  \includegraphics[width=\linewidth]{
    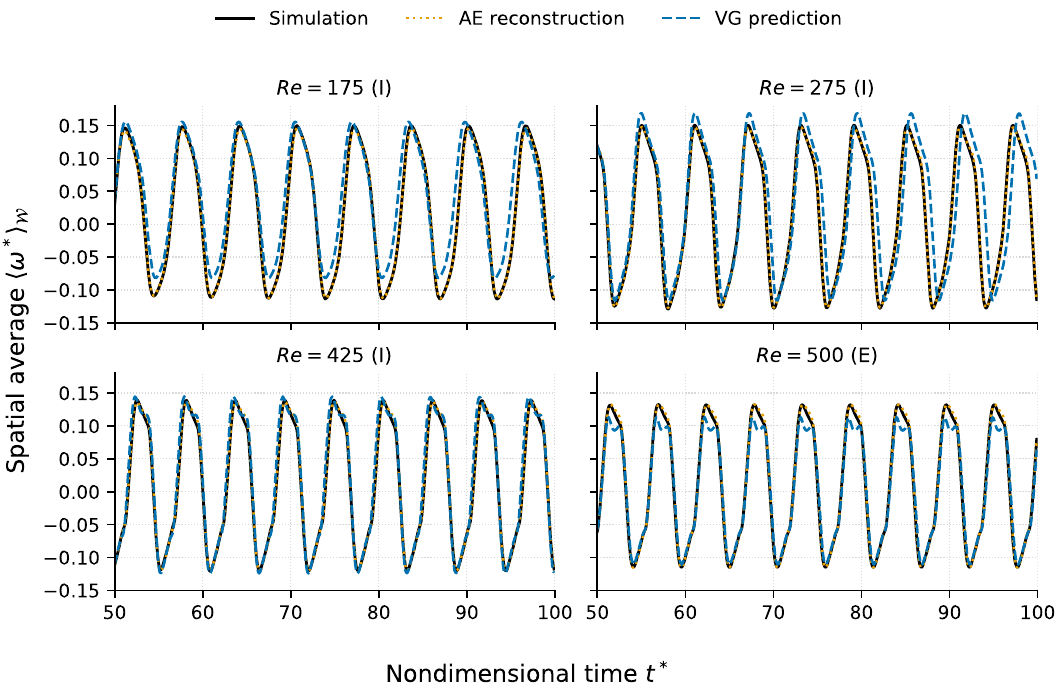}
  \caption{Spatially averaged wake vorticity from the simulation,
  autoencoder reconstruction, and post-optimization VG prediction. I denotes
  interpolation within the symbolic model-construction range; E denotes
  extrapolation beyond that range.}
  \label{fig:supp-multi-re-spatial-average}
\end{figure}

The autoencoder traces in
Figure~\ref{fig:supp-multi-re-spatial-average} closely follow the simulation
at all four Reynolds numbers, indicating that the common representation
preserves this integral wake quantity. The VG traces retain the
oscillation frequency and amplitude range throughout each withheld interval.
Agreement is strongest at $\Rey=425$ and $500$. At $\Rey=175$ the predicted
phase gradually separates from the reconstructed field, while
$\Rey=275$ exhibits the largest late-time phase difference, consistent with
its larger latent error. All four cases remain within the periodic
vortex-shedding regime, whose dominant oscillator provides a common
dynamical structure across Reynolds number. The residual differences are
consistent with the selected Reynolds-conditioned equation capturing this
shared oscillator more accurately than the detailed variation of its phase
and amplitude.

\subsection{Field reconstruction and error decomposition}

The field comparison uses the same three-part error decomposition as
Eq.~\eqref{eq:supp-fixed-field-errors}. Here the POD reconstruction
$\boldsymbol{\omega}_{\mathrm{POD}}$ is replaced by the autoencoder
reconstruction $\boldsymbol{\omega}_{\mathrm{AE}}$. Accordingly,
$e_{\mathrm{AE}}$ and $e_{\mathrm{VG\text{-}AE}}$ replace
$e_{\mathrm{POD}}$ and $e_{\mathrm{VG\text{-}POD}}$, respectively, while the
end-to-end definition is unchanged.

Table~\ref{tab:supp-multi-re-errors} summarizes the latent and field errors
over all 500 snapshots. The mean autoencoder error varies only from 5.90\%
to 6.96\% across the four cases, whereas the VG-to-autoencoder error varies
more substantially. The difference in end-to-end accuracy is therefore
associated mainly with the symbolic rollout rather than a deterioration of
the common field representation. As in the fixed-$\Rey$ decomposition, the
end-to-end error is not the arithmetic sum of its two components.

\begin{table}[!ht]
  \centering
  \caption{Extended cross-$\Rey$ latent and field errors. Field entries give
  the temporal mean with the median in parentheses, in percent.}
  \label{tab:supp-multi-re-errors}
  \small
  \setlength{\tabcolsep}{3.2pt}
  \begin{tabular}{llccccc}
    \toprule
    $\Rey$ & Regime &
    \multicolumn{2}{c}{Latent NMSE} &
    AE error & VG--AE error & End-to-end error \\
    \cmidrule(lr){3-4}
    & & Pre-opt. & Post-opt. & Mean (median) & Mean (median) & Mean (median) \\
    \midrule
    175 & Interpolation & 2.312 & 0.142 & 6.96 (6.81) & 19.85 (18.30) & 20.86 (19.13) \\
    275 & Interpolation & 0.797 & 0.247 & 5.90 (5.83) & 32.07 (31.44) & 33.01 (32.50) \\
    425 & Interpolation & 0.188 & 0.043 & 6.00 (5.98) & 18.88 (19.97) & 20.74 (21.72) \\
    500 & Extrapolation & 0.714 & 0.036 & 6.41 (6.46) & 11.89 (13.10) & 13.81 (14.26) \\
    \bottomrule
  \end{tabular}
\end{table}

\begin{figure}[p]
  \centering
  \includegraphics[width=\linewidth]{
    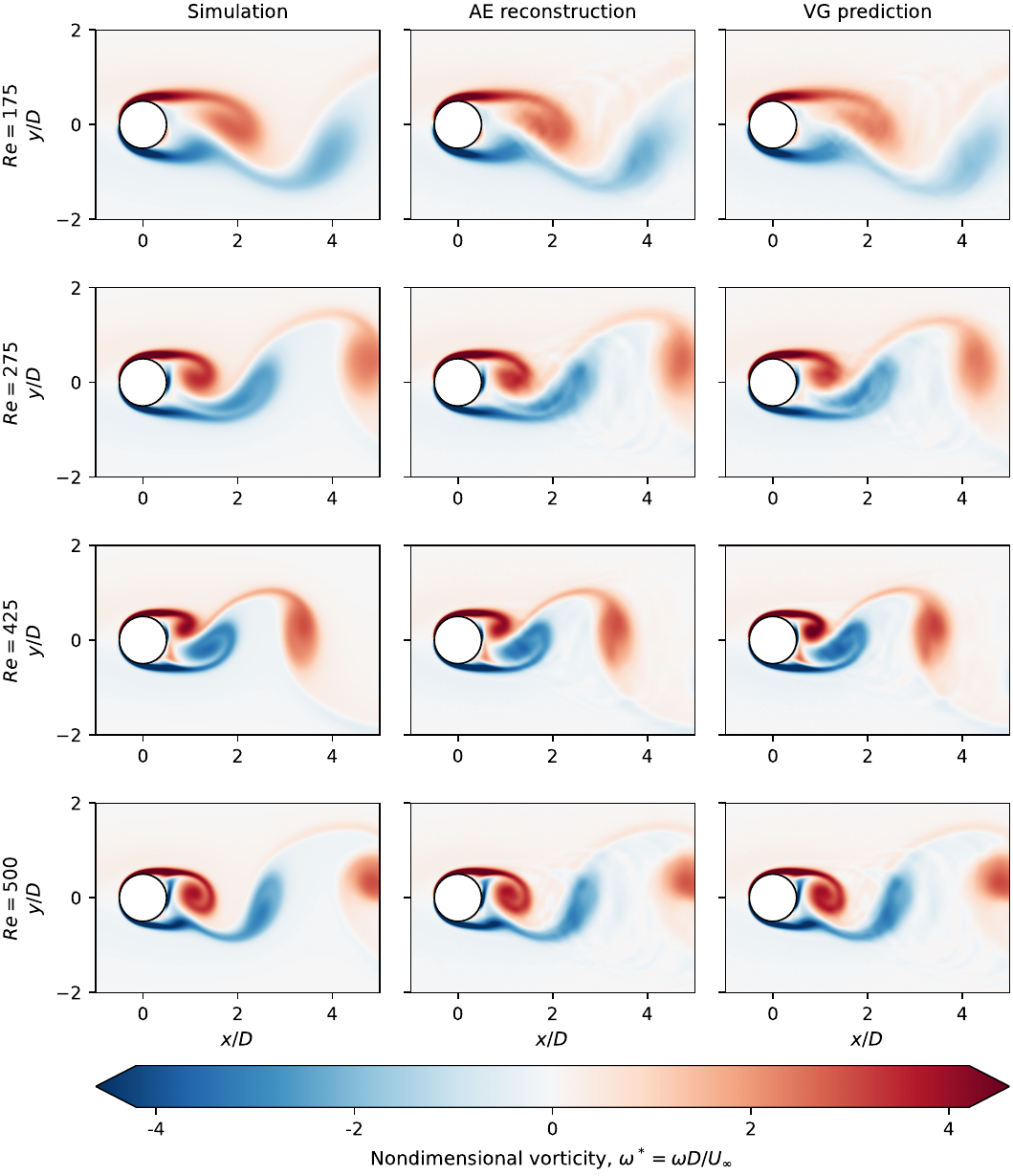}
  \caption{Simulation, autoencoder reconstruction, and VG prediction at
  $t^*=75$ for all Reynolds numbers excluded from symbolic model
  construction. The first three rows are interpolation cases; the final row
  is extrapolation of the symbolic dynamics.}
  \label{fig:supp-multi-re-fields}
\end{figure}

Figure~\ref{fig:supp-multi-re-fields} consolidates all four withheld Reynolds
numbers at $t^*=75$, including the $\Rey=275$ and $500$ cases presented in the
paper, so that the field predictions can be compared directly. The
autoencoder consistently smooths smaller spatial features while preserving
the alternating wake. The VG prediction reconstructs the dominant structures
available in the three-coordinate representation at every Reynolds number.
Its greater displacement from the encoded field at $\Rey=275$ is consistent
with the accumulated phase difference in
Figures~\ref{fig:supp-multi-re-latent} and
\ref{fig:supp-multi-re-spatial-average}; the closer latent rollouts at
$\Rey=425$ and $500$ correspond to more closely aligned large-scale fields.

As discussed in Section~\ref{sec:supp-multi-re-latent}, the recovered
equation represents a Reynolds-conditioned common oscillator rather than a
complete symbolic description of all parameter-dependent wake dynamics.
Even with this qualification, obtaining one explicit system that remains
bounded and predictive at four Reynolds numbers excluded from symbolic model
construction---including one outside the construction range---is nontrivial
for reduced coordinates of a high-dimensional flow, particularly when using
a synthetically pretrained symbolic transformer as the backbone. Parametric
symbolic modeling of physical flow systems remains an open challenge; these
results demonstrate the value of combining shared oscillatory structure,
explicit parameter dependence, and cross-trajectory verification.

\FloatBarrier

\end{document}